\pdfoutput=1

\documentclass[11pt]{article}
\usepackage{kotex}
\usepackage{booktabs}
\usepackage[preprint]{acl}

\usepackage{times}
\usepackage{latexsym}

\usepackage[T1]{fontenc}

\usepackage[utf8]{inputenc}

\usepackage{microtype}

\usepackage{inconsolata}

\usepackage{graphicx}

\usepackage{booktabs}       %
\usepackage{array}
\usepackage[
  separate-uncertainty = true,
  multi-part-units = repeat
]{siunitx}
\usepackage{multirow}
\usepackage{graphicx}
\usepackage{relsize}
\usepackage{amsmath,amssymb}
\usepackage{tabularx} %
\usepackage{lipsum} %
\usepackage{float}
\usepackage{placeins}

\title{KMI: A Dataset of Korean Motivational \\Interviewing Dialogues for Psychotherapy}

\author{
    \textbf{Hyunjong Kim}$^{1}$ \quad
    \textbf{Suyeon Lee}$^{1}$\thanks{Equal contribution.} \quad
    \textbf{Yeongjae Cho}$^{1}$\footnotemark[1] \\
    \textbf{Eunseo Ryu}$^{1}$ \quad
    \textbf{Yohan Jo}$^{1}$ \quad
    \textbf{Suran Seong}$^{2}$ \quad
    \textbf{Sungzoon Cho}$^{1}$\thanks{Corresponding author.} \\
    $^1$Seoul National University \quad
    $^2$Korea Counseling Graduate University \\
    \texttt{\{hjkim, suyeon.lee, yjcho\}@bdai.snu.ac.kr} \\
    \texttt{\{eunseoryu, yohan.jo, zoon\}@snu.ac.kr} \\
    \texttt{nina9698@gmail.com}
}

\begin{document}
\maketitle
\begin{abstract}
The increasing demand for mental health services has led to the rise of AI-driven mental health chatbots, though challenges related to privacy, data collection, and expertise persist. Motivational Interviewing (MI) is gaining attention as a theoretical basis for boosting expertise in the development of these chatbots. 
However, existing datasets are showing limitations for training chatbots, leading to a substantial demand for publicly available resources in the field of MI and psychotherapy. These challenges are even more pronounced in non-English languages, where they receive less attention.
In this paper, we propose a novel framework that simulates MI sessions enriched with the expertise of professional therapists. We train an MI forecaster model that mimics the behavioral choices of professional therapists and employ Large Language Models (LLMs) to generate utterances through prompt engineering.
Then, we present \textbf{KMI}, the first synthetic dataset theoretically grounded in MI, containing 1,000 high-quality \textbf{K}orean \textbf{M}otivational \textbf{I}nterviewing dialogues.
Through an extensive expert evaluation of the generated dataset and the dialogue model trained on it, we demonstrate the quality, expertise, and practicality of KMI. We also introduce novel metrics derived from MI theory in order to evaluate dialogues from the perspective of MI.
\end{abstract}

\section{Introduction}
\label{sec:intro}

In modern society, the issue of mental health is emerging as a critical problem, with an increasing demand for mental health services. This has led to a shortage of mental health workers available to meet the growing demand \citep{butryn2017shortage}. Thus, there has been considerable expansion in the investigation of AI-driven chatbots providing mental health assistance \citep{inkster2018empathy, mousavi2021would}. However, there are many challenges in training such chatbots. It suffers from privacy issues, requires a considerable amount of cost and time for collecting data, and crucially, needs expertise in mental health.
\citet{cho2023integrative} pointed out that many mental health chatbots are designed without any underlying theory for psychotherapy, casting doubt on their utility for mental health support.

\begin{figure}[t!]
    \centering 
    \includegraphics[width=1.0\linewidth]{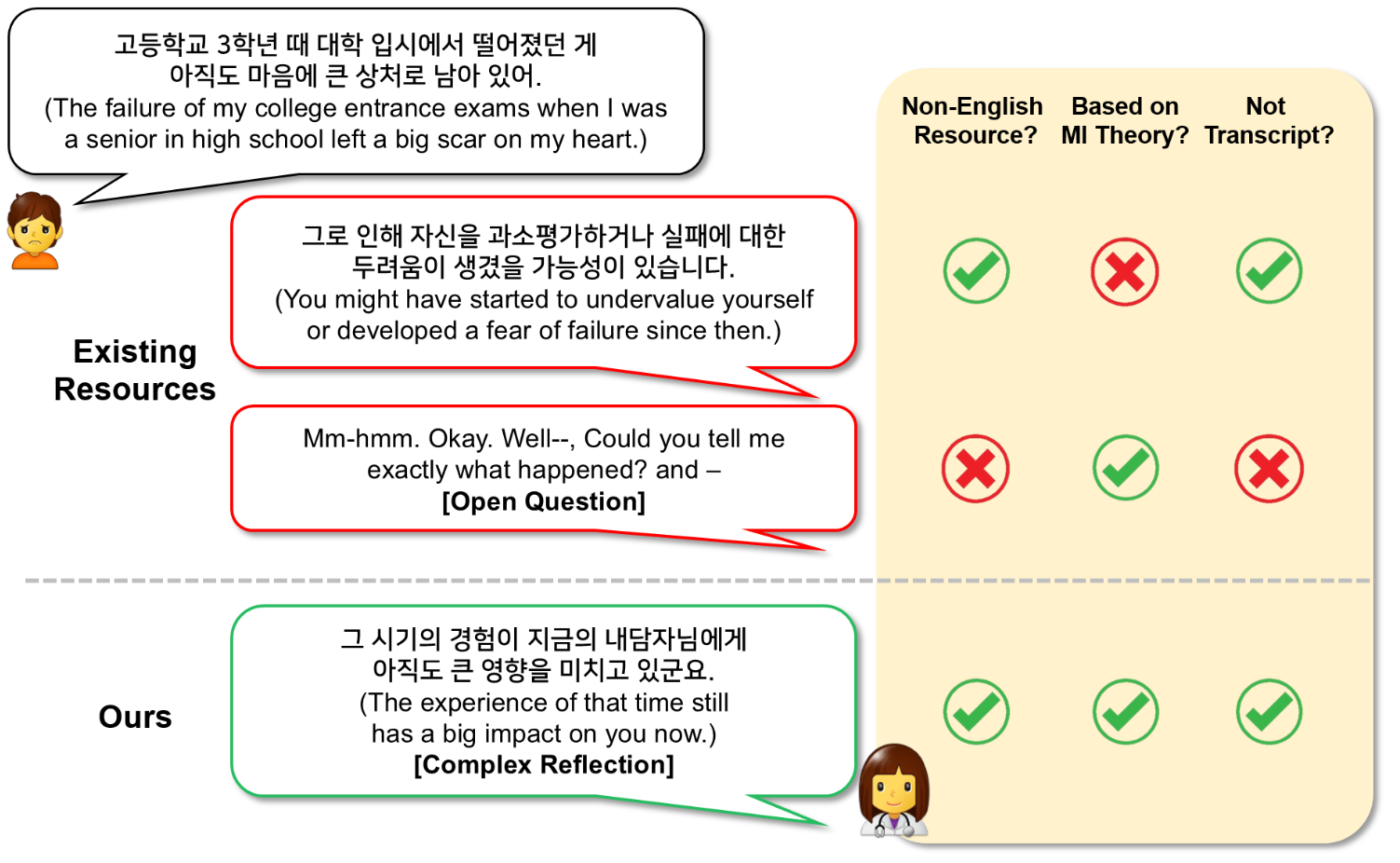}
    \caption{Comparison between KMI and existing resources. KMI is the only dataset that is a non-English resource, based on MI theory, and not derived from transcripts. Other datasets fall short in some of these aspects, limiting their suitability for chatbot applications.} 
    \label{fig:intro}
\end{figure}

As a theoretical basis for enhancing expertise, Motivational Interviewing (MI) has been attracting increasing attention in developing mental health support chatbots \citep{park2019designing, he2022can, brown2023motivational}. 
MI is a client-centered counseling technique to elicit behavior change by helping clients explore and resolve ambivalence \citep{miller2002motivational}. MI has been an active area of research intersecting the disciplines of psychotherapy and Natural Language Processing (NLP), due to its effectiveness and well-structured behavioral coding systems such as MITI (Motivational Interviewing Treatment Integrity) \citep{moyers2003motivational, moyers2014motivational}. 

Few works have proposed MI datasets with annotations from MI behavioral coding systems. 
\citet{perez2016building} developed an MI dataset using automatic captioning of YouTube and Vimeo videos. However, the dataset is currently not publicly available due to ethical reasons.
\citet{welivita2022curating} proposed an alternative of scraping data from online peer support forums. Although these dialogues are available in abundance, they have shown significant differences from those collected from professional counselors in terms of MI, thus lacking expertise.
In line with this problem, \citet{wu2023creation} presented \textit{AnnoMI}, a dataset of 133 MI conversations that were professionally transcribed from MI demonstration videos and further annotated by experienced MI practitioners. Despite its decent transcription and annotation quality, it significantly lacks in quantity due to the limited video sources and the intensive human labor required.

In short, the mentioned datasets contain limitations to be used for training MI-boosted chatbots, such as lack of expertise or insufficient quantity.
Also, transcripts often include many onomatopoeic words and instances of stammering, which are undesirable in chatbots.
Furthermore, previous works have predominantly focused on English. While increasing language coverage is crucial in addressing the challenges of mental health support \citep{cho2023integrative}, research on other languages and cultures has been relatively understudied.

To address these issues, we propose a novel framework to generate synthetic dialogues that simulate MI sessions using Large Language Models (LLMs).
We carefully design two agents, the therapist simulator and the client simulator, which alternately take turns to generate utterances based on few-shot in-context learning.
To incorporate the expertise of professional therapists, we train an MI forecaster model, which predicts the next-turn therapist behavior.
Also, we focus on a non-English language, Korean in our case\footnote{South Korea is struggling with widespread mental health problems such as depression and anxiety, highlighted by the highest suicide rate among OECD countries \citep{/content/publication/7a7afb35-en}.}, and ground the dialogues in real-world contexts that reflect actual Korean circumstances.
Consequently, we present \textbf{KMI}, a dataset of 1,000 \textbf{K}orean \textbf{M}otivational \textbf{I}nterviewing dialogues. 
As illustrated in Figure \ref{fig:intro}, KMI is the only dataset that overcomes the previously mentioned limitations of existing resources.
KMI covers a wide range of concerns and anxieties common among Koreans, while successfully integrating the MI strategy of professional therapists simultaneously. Each therapist utterance within the dataset is annotated with a therapist behavior label.
To the best of our knowledge, our work is the first attempt to construct an MI dataset using the generative capability of LLMs.

We evaluate KMI through a comprehensive evaluation process by engaging professional counselors. We first validate the dataset's quality from the perspective of both MI and general dialogue systems. Notably, for the MI-driven evaluation, we introduce novel metrics derived from MI theory to directly measure how well the conversation complies with the spirit of MI. Then, we train a dialogue model using our dataset to assess its utility as a resource for training a chatbot for mental health support.
Evaluation results demonstrate that our dataset not only effectively captures the essence of MI, but also offers practical utility for chatbot development.

Our contributions are as follows.
\begin{enumerate}
    \item We propose a novel framework that simulates MI sessions enriched with the expertise of professional therapists. This is achieved by training and leveraging an MI forecaster model that mimics the behavioral choices of real-world therapists.
    \item We present KMI, the first synthetic dataset theoretically grounded in MI, which contains 1,000 high-quality Korean MI dialogues. We publicly release our dataset to address the issue of data shortage in psychotherapy\footnote{\url{https://github.com/hjkim811/KMI}}.
    \item Through an extensive expert evaluation of the generated dataset and the dialogue model trained on it, we demonstrate the quality, expertise, and practicality of KMI. We also introduce novel metrics derived from MI theory in order to evaluate dialogues from the perspective of MI.
\end{enumerate}

\section{Related Work}
\label{sec:related_work}

\begin{figure*}[htb!]
    \centering 
    \includegraphics[width=1\linewidth]{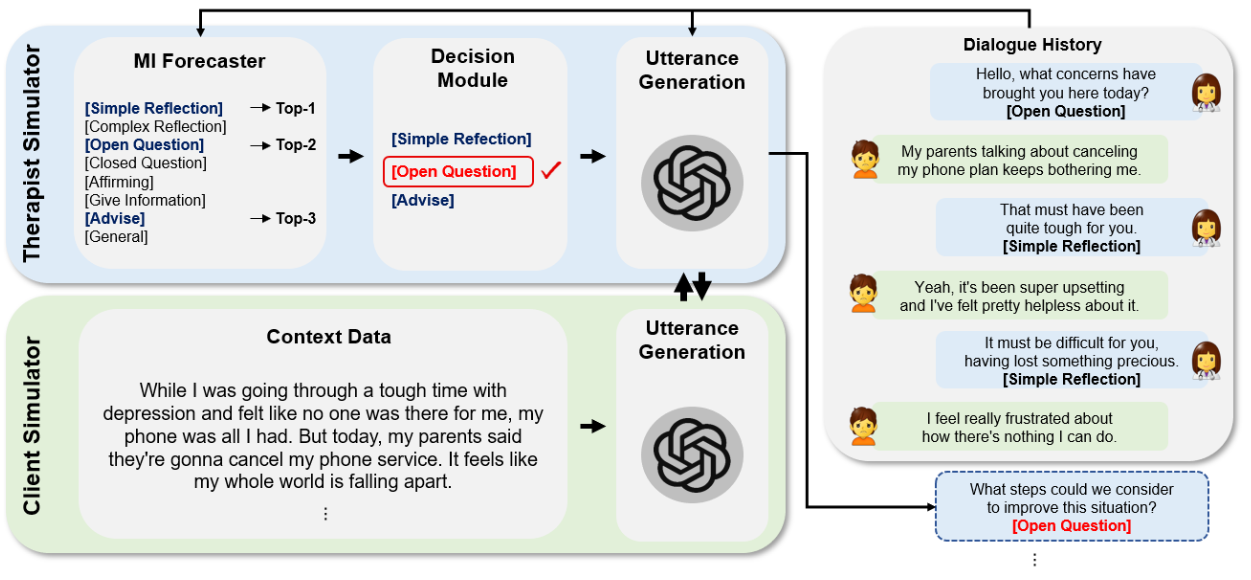}
    \caption{The overall framework for generating the KMI dataset. The context data and dialogue history are originally in Korean but have been translated into English for the figure.} 
    \label{fig:framework}
\end{figure*}

\paragraph{Dialogue Generation Using LLMs}
Recent research has increasingly focused on using LLMs to generate dialogues for various applications.
\citet{kim2022prosocialdialog} proposed a human-machine collaborative framework to build a large-scale dialogue dataset for training conversational agents to handle problematic content appropriately.
\citet{chen2023places} used a small number of expert-written conversations as in-context examples to create synthetic multi-party conversations.
\citet{chen2023controllable} proposed LLM prompting methods to generate mixed-initiative dialogues.
\citet{kim2023soda} created a large-scale social dialogue dataset by distilling conversations from LLMs.
\citet{macina2023mathdial} paired human teachers with an LLM student to generate teacher-student tutoring dialogues grounded in math reasoning problems.

\paragraph{NLP Applications in MI}
MI was developed as a technique to assist individuals in resolving ambivalence and committing to change \citep{Miller_1983}, representing an evolution of client-centered therapy. 
Along with the advancement in NLP, considerable research efforts have been underway to apply NLP techniques in the field of MI. 
Several studies have been proposed to automatically classify a given utterance into one of the MI behavioral codes.
While earlier approaches utilize linguistic features \citep{perez2017predicting} and recurrent neural network architectures \citep{tanana2015recursive, xiao2016behavioral, cao2019observing, gibson2019multi}, recent approaches make use of pre-trained language models such as RoBERTa \citep{liu2019roberta, tavabi2021analysis, welivita2023boosting}. 
Some works adopt a multimodal approach, leveraging additional information such as speech features \citep{tavabi2020multimodal} or facial features \citep{nakano2022detecting}.

\citet{welivita2023boosting} demonstrated the potential of LLMs in boosting dialogues using the MI strategy. They observed that among the MI dataset curated from online platforms \citep{welivita2022curating}, 92.86\% of the advice given by peers falls into the \textit{Advise without permission} category, which is MI non-adherent. To make the dataset more MI-consistent, they fine-tuned BlenderBot \citep{roller2021recipes} and GPT-3 \citep{brown2020language} to rephrase these responses into more MI-adherent \textit{Advise with permission} responses. Although this work demonstrated the possibility of leveraging LLMs in MI, the impact of rephrasing remains marginal, considering that its capability is restricted to revising the manner of speech and has limitations in modifying the content itself.
In our work, we further exploit the generative ability of LLMs to generate the entire dataset from scratch.

\begin{table*}[hbt!]
    \begin{center}
    \scriptsize
    \setlength{\tabcolsep}{10pt}
    \renewcommand{\arraystretch}{1.2}
    \begin{tabular}{p{2.2cm}p{6cm}p{5.7cm}}
    \toprule
    \textbf{MI Label} & \textbf{Description}  & \textbf{Examples} \\ 
    \midrule
    1. Simple Reflection & Repetition, rephrasing, or paraphrasing of the speaker’s previous statement. & It sounds like you’re feeling worried. \\
    2. Complex Reflection & Repeating or rephrasing the previous statement of the speaker but adding substantial meaning/emphasis to it. & \parbox[t]{5.6cm}{\textit{Speaker}: Mostly, I would change for future generations.\\ \textit{Listener}: It sounds like you have a strong feeling of responsibility.} \\
    3. Open Question & Questions that allow a wide range of possible answers. & What is your take on that? \\
    4. Closed Question & Questions that can be answered with a yes/no response or a very restricted range of answers. & Do you think this is an advantage? \\
    5. Affirm & Encouraging the speaker by saying something positive or complimentary. & You should be proud of yourself for your past efforts. \\
    6. Give Information & Educating, providing feedback, or giving an opinion without advising. & Logging your cravings is important as cravings often lead to relapses. \\
    7. Advise & Making suggestions, offering solutions or possible actions. & We could try to brainstorm some ideas that might help. \\
    8. Other & Statements that are not classified under the above codes. & Hi there.
     \\
    \bottomrule
    \end{tabular}
    \end{center}
    \caption{MI labels derived from the MITI code. The descriptions and examples of each label are taken from \citet{welivita2022curating}.}
    \label{tab:mi_label}
\end{table*}

\section{KMI: A Dataset of Korean Motivational Interviewing Dialogues}
\label{sec:method}

This section introduces a novel framework to generate realistic motivational interviewing dialogues. The overall framework is illustrated in Figure \ref{fig:framework}. First, we collect context data (Section \ref{sec:context}) that reflects the actual concerns and anxieties experienced by Koreans. Each context serves as the topic of each dialogue. Next, we simulate an MI session (Section \ref{sec:mi_simulation}) using a therapist simulator and client simulator. The resulting KMI dataset is presented in Section \ref{sec:results}.

\subsection{Collecting Context Data}
\label{sec:context}
To construct context data for generating realistic dialogues, we web-crawl posts from a Korean psychological counseling platform \textit{Mindcafe}\footnote{\url{https://www.mindcafe.co.kr/pc/community}}, which contain common concerns of Koreans. We collect a total of 7,530 posts from seven categories: mental health, interpersonal relationships, ego \& personality, career \& employment, academic \& examination, addiction \& obsession, and family. 
Then, using GPT-4 API\footnote{\texttt{gpt-4-0125-preview}} \citep{achiam2023gpt}, we score these posts on a scale from 1 to 3 in terms of the post's specificity and its suitability as a topic for an MI session.
Among the 3,098 posts that received a score of 3, a total of 1,000 posts were sampled with a predetermined quantity for each category.
Details regarding the collection, filtering, and sampling of context data can be found in Appendix~\ref{sec:context_data}.

\subsection{Motivational Interviewing} %
\label{sec:mi_simulation}
For each context, we simulate an MI session using an LLM-based therapist simulator (Section \ref{sec:therapist_simulator}) and client simulator (Section \ref{sec:client_simulator}). They alternately take turns to generate utterances. The conversation begins with a general open question by the therapist, such as `Hello, what concerns have brought you here today?'. For both simulators, we utilize GPT-4 API for high-quality generation. 

We define eight labels derived from MITI code 2.0 \citep{moyers2003motivational} and 4.2.1 \citep{moyers2014motivational} to categorize therapist behaviors. Table \ref{tab:mi_label} shows the description and example of each label.
Each therapist utterance in the final KMI dataset is annotated with one of these eight labels, which helps capture the nuances of therapeutic conversations and enhances the dataset's utility for various NLP applications.

\subsubsection{Therapist Simulator}
\label{sec:therapist_simulator}
Using MI-consistent techniques and blending them skillfully helps motivate clients to change \citep{moyers2014motivational}. 
Thus, it is crucial to follow the behavioral choices of professional therapists to generate high-quality MI dialogues. In this work, this is achieved by first predicting the next-turn therapist behavior via MI forecaster and decision module, and then generating the utterance based on the predicted behavior by prompting an LLM.
If one could simulate a realistic therapist with expertise, LLM-based simulation could be a significant breakthrough in the field of MI, alleviating the issues of extensive human workload and privacy concerns.

\paragraph{MI Forecaster}
MI forecaster aims to predict the next-turn therapist behavior, which is one of the MI labels listed in Table \ref{tab:mi_label}, given a dialogue history. We fine-tune T5-base \citep{raffel2020exploring} with a converted dataset of AnnoMI \citep{wu2023creation}, which is preprocessed and converted for a forecasting task. AnnoMI is an expert-annotated MI dialogue dataset. As the original dataset doesn't include \textit{Affirm} among its utterance labels, we add \textit{Affirm} label to the dataset using the RoBERTa-based MI classifier\footnote{The checkpoints can be found in \url{https://github.com/anuradha1992/Boosting-with-MI-Strategy}.} developed in \citet{welivita2023boosting}. More details regarding the preprocessing of AnnoMI can be found in Appendix \ref{sec:annomi_preprocessing}.

Then we construct pairs of input text and output text, where input text is the dialogue history prefixed with the task instruction (\texttt{Predict next therapist's dialogue act}) and output text is the next-turn therapist behavior label. We explore various modeling settings to find the model with the best predictive power: (1) we test history window sizes from 1 to 8 and (2) try inserting the MI behavior label of each therapist utterance in the dialogue history to provide additional information to the model.
An example of the converted dataset is shown in Table \ref{tab:t5_dataset_ex} of Appendix \ref{sec:mi_forecaster_details}. To form the input text, we concatenate the utterances in dialogue history, each prefixed with a special token indicating the speaker. In the setting where therapist labels are provided, the annotated label of every therapist utterance in dialogue history is inserted in the special token.

Considering the relatively small size of the dataset, we use 5-fold cross-validation to evaluate each model setting. Statistics of the dataset and implementation details can be found in Appendix \ref{sec:mi_forecaster_details}.
Though the next-turn label from AnnoMI is used as the ground-truth label, it is not the only correct choice given the nature of the task; even in the same situation, different therapists might opt for different strategies and approaches to deal with the client depending on their counseling style and therapeutic philosophy. Therefore, we use top-3 accuracy instead of top-1 accuracy when monitoring the model performance, assuming that the response is plausible if the model's prediction matches the label in the dataset within three attempts. 

\begin{figure}[t!]
    \centering 
    \includegraphics[width=1.0\linewidth]{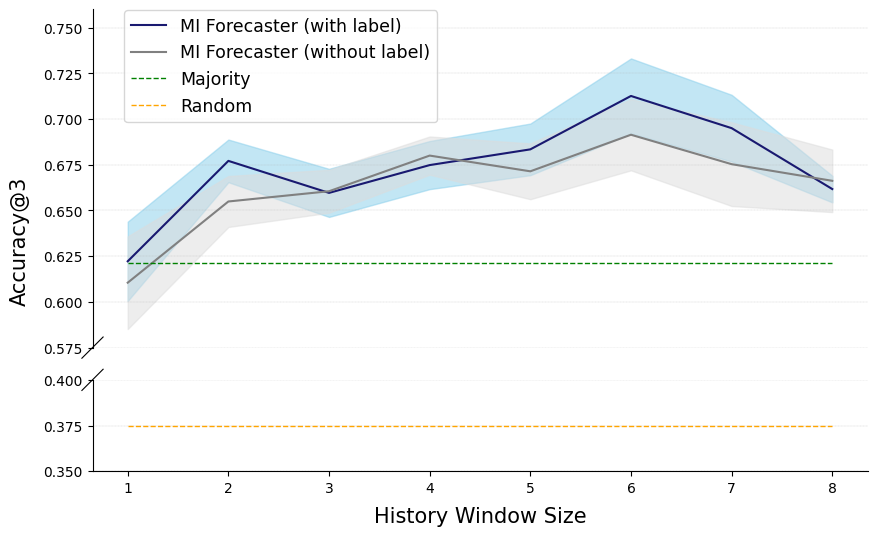}
    \caption{Top-3 accuracy of the MI forecaster using 5-fold cross-validation. The bold line represents the mean accuracy across folds and the shaded area indicates the 95\% confidence interval.} 
    \label{fig:mi_forecaster_results}
\end{figure}

The experimental results are illustrated in Figure \ref{fig:mi_forecaster_results}. We observe that the model performs best when six utterances are given as dialogue history and therapist labels are provided, demonstrating 71.26\% top-3 accuracy.
Also, inserting therapist labels in the dialogue history has improved accuracy and reduced cross-fold variance, offering benefits in the forecasting task.
To verify the performance of the MI forecaster, we provide additional baselines \textit{Majority} and \textit{Random}. Majority predicts the three most frequent labels for top-3 prediction, while Random predicts labels randomly. Our model significantly outperforms these baselines, validating its effectiveness.

Finally, we train the final MI forecaster model for the generation process of KMI. Based on the experimental results, we adopt the optimal setting previously mentioned and use all available data for training, as a held-out test set is no longer required at this stage. In each therapist's turn, the MI forecaster predicts the three most likely MI labels based on the previously generated dialogue history. Since the generated utterances that function as input text are in Korean, we first translate the dialogue history into English and then concatenate six recent utterances to form the input text. We utilize Upstage SOLAR \citep{kim-etal-2024-solar} translation API\footnote{\texttt{solar-1-mini-translate-koen} model} for translation.

\paragraph{Decision Module}
In each therapist's turn, among the three labels the MI forecaster predicted, the decision module decides the final therapist label to generate. Based on two simple rules, it complements the MI forecaster from a broader perspective: (1) The same therapist label cannot appear three times in a row. (2) Either open or closed, the therapist cannot ask questions three times in a row. The first rule is devised to keep the dialogue from being too homogeneous. The second rule is a direct implementation of a clinical guideline that advises against asking three questions in a row \citep{miller2002motivational}. Such behavior might direct the client into a passive, question-answering role, which should be avoided in MI.
In the order of top-1, top-2, and top-3 predictions, the decision module checks each label against these two rules. The first prediction that complies with both rules is determined to be the final therapist's behavior for the next turn.

\paragraph{Utterance Generation}
Once the label is determined, we generate the therapist's utterance by prompting an LLM to generate an utterance based on the determined label and dialogue history. We leverage in-context learning for this purpose, providing the definition and three examples of the corresponding label. The definition and examples are excerpted from Korean MI textbooks \citep{korean_mi_textbook_2017, korean_mi_textbook_2016, korean_mi_textbook_2019}, which could be considered among the most credible sources available. These examples contain speech patterns and linguistic expressions actually used by professional Korean counselors. In all prompts, we specify predefined constraints to generate consistent and natural utterances. Also, the simulator is instructed to conclude the conversation if the client's concerns seem to have been resolved.
The prompt for generating a simple reflection utterance, which includes instructions, constraints, definitions, and examples of simple reflection, and dialogue history is shown in Table \ref{tab:prompt_table_1} of Appendix \ref{sec:prompt_templates}.
Along with the generated utterances, the corresponding MI labels for each utterance are included in our dataset as utterance-level annotations.

\begin{table}[t!]
    \small
    \begin{center}
    \renewcommand{\arraystretch}{1}
    \resizebox{\columnwidth}{!}{%
    \begin{tabular}{lrrr}
    \toprule
    \textbf{Dialogue Statistics} & Total & Therapist & Client   \\ \midrule
    \# Dialogues  & 1,000   & -      & -                        \\
    \# Turns      & 18,116  & 9,558  & 8,558                    \\
    Avg. Turns per Dialogue & 18.12  & 9.56       & 8.56        \\ \midrule
    \multicolumn{4}{l}{\textbf{\# Therapist MI Label}}  \\ \midrule
    Simple Reflection       & \multicolumn{3}{r}{1,269 (15\%)}  \\
    Complex Reflection      & \multicolumn{3}{r}{3,055 (36\%)} \\
    Open Question           & \multicolumn{3}{r}{2,305 (27\%)}  \\
    Closed Question         & \multicolumn{3}{r}{109 (1\%)}     \\
    Affirm                  & \multicolumn{3}{r}{914 (11\%)}    \\
    Give Information        & \multicolumn{3}{r}{87 (1\%)}      \\
    Advise                  & \multicolumn{3}{r}{43 (1\%)}      \\
    Other                   & \multicolumn{3}{r}{779 (9\%)}     \\ \bottomrule
    \end{tabular}%
    }
    \end{center}
    \caption{Statistics of KMI.}
    \label{tab:2}
\end{table}

\subsubsection{Client Simulator}
\label{sec:client_simulator}
We generate client utterances by directly prompting an LLM. We provide the context and instruct the LLM to generate responses based on the context and dialogue history. In order to depict the evolving states of clients throughout the MI session, we adopt a simple but effective approach to instruct the LLM to generate \textit{change talk}, a client language that indicates movement toward a particular change \citep{miller2023motivational}, if interaction with the counselor inspired a willingness to change and speaking about change appears suitable within the dialogue context.
As in the therapist simulator, we provide the definition and four examples of change talk excerpted from Korean MI textbooks \citep{korean_mi_textbook_2017, korean_mi_textbook_2016, korean_mi_textbook_2019} for in-context learning. The four examples each include one from the four types of preparatory change talk—desire, ability, reasons, and need (often abbreviated as DARN) \citep{miller2023motivational}—encompassing various types of change talk. Descriptions of each type can be found in Appendix \ref{sec:change_talk}. Also, we specify predefined constraints in each prompt. The prompt for generating a client utterance, which includes instructions, constraints, definition, and examples of change talk, context, and dialogue history is shown in Table \ref{tab:prompt_table_2} of Appendix \ref{sec:prompt_templates}.
Based on the proposed framework, we finally generate KMI, a dataset consisting of 1,000 Korean MI dialogues.

\subsection{Statistics}
\label{sec:results}
\begin{figure}[t!]
    \centering 
    \includegraphics[width=1.0\linewidth]{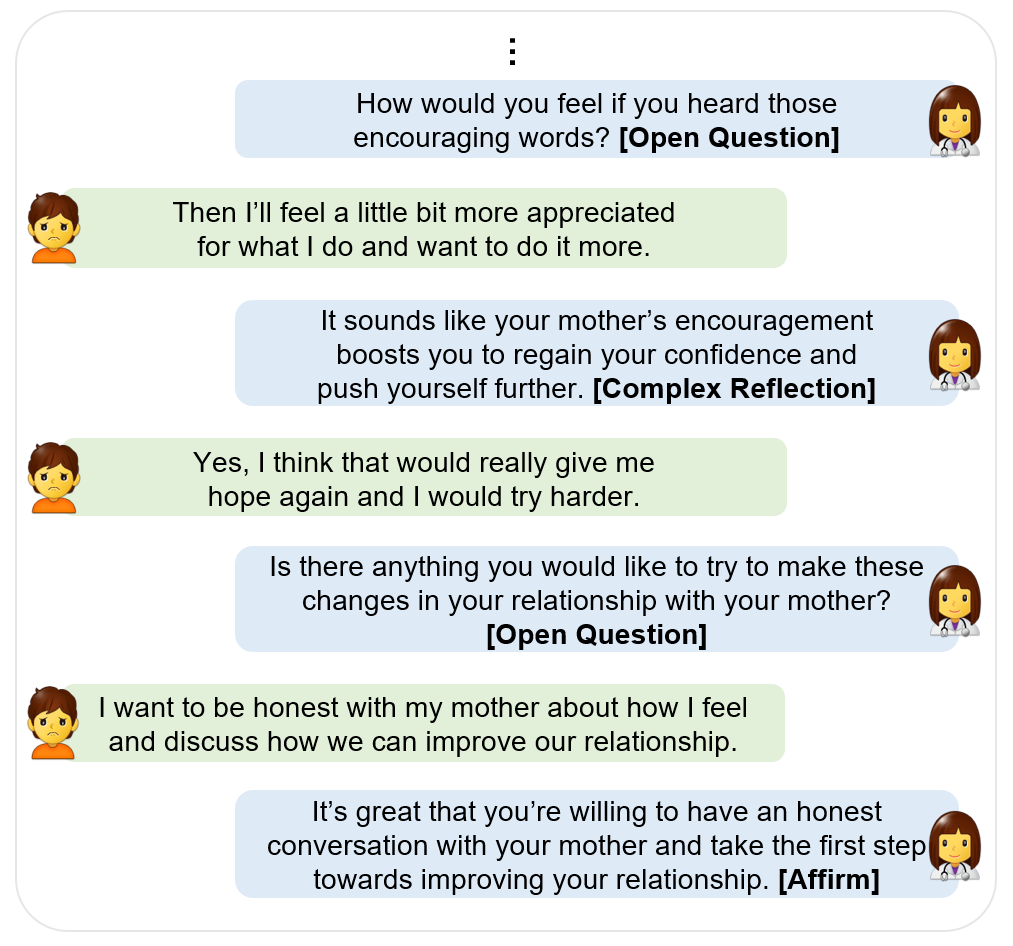}
    \caption{An example of KMI dataset. The dialogue has been translated into English for the figure. Examples of full dialogues are provided in Appendix \ref{sec:dialogue_examples_kmi}.} 
    \label{fig:examples}
\end{figure}

As shown in Table \ref{tab:2}, KMI consists of 1,000 long-turn dialogues, with an average of 18.12 turns per dialogue.
Each utterance by therapists is assigned one of the MI labels.
As detailed in Table \ref{tab:2}, \textit{Complex Reflection} emerges as the most frequent label within KMI, accounting for 36\% of all therapist utterances, followed by \textit{Open Question} and \textit{Simple Reflection}. 
Although there are no strict guidelines for the proportion of MI labels, MITI coding manual \citep{moyers2014motivational} provides summary scores to measure clinicians' competence in using MI. A Reflection-to-Question Ratio (R:Q) of 1:1 is considered ‘fair’, while 2:1 is considered ‘good’. KMI yields a ratio of 1.8:1, implying that it meets the standard of professional clinicians.
An example dialogue in KMI is illustrated in Figure \ref{fig:examples}.

\section{Evaluation}
\label{sec:evaluation}

We evaluate the quality of the generated dataset (Section \ref{sec:evaluation_dataset}) and the dialogue model fine-tuned with it  (Section \ref{sec:evaluation_chatbot}).
We compare our dataset with CounselGPT\footnote{\url{https://github.com/MrBananaHuman/CounselGPT}}, the only Korean counseling dataset to date that is created using OpenAI API, and AnnoMI. We translate AnnoMI into Korean using Upstage SOLAR \citep{kim-etal-2024-solar} translation API\footnote{\texttt{solar-1-mini-translate-enko} model}, in order to evaluate it in the context of Korean.

\subsection{Evaluation Criteria}
\label{sec:evaluation_criteria}
We evaluate the datasets and dialogue models from two perspectives: MI and general dialogue systems.

\paragraph{MI Quality}
We aim to measure how closely the dialogues adhere to the principles of MI. To achieve this goal, we propose novel metrics derived from MI theory: (1) \textbf{Partnership}, (2) \textbf{Acceptance}, (3) \textbf{Compassion}, (4) \textbf{Evocation}, (5) \textbf{Similarity}, and (6) \textbf{Effectiveness}. (1)\textasciitilde(4) are derived from the fundamental spirit of MI~\citep{miller2012motivational}. 
Similarity measures how closely the generated therapist's utterances resemble those of an actual therapist.
Effectiveness measures the overall efficacy of the MI session.
Descriptions of each evaluation criterion can be found in Appendix~\ref{sec:eval_criteria}.

\paragraph{General Quality}
Following previous works \citep{wan2022unified, chen2023places}, we use three evaluation criteria to measure the general quality of the dialogues: (1) \textbf{Consistency}, (2) \textbf{Fluency}, and (3) \textbf{On-topic}. Consistency examines whether the entire dialogue and utterances between turns are consistent. Fluency assesses the flow of the dialogue and the naturalness and fluidity of each utterance. On-topic evaluates whether the dialogue is relevant to the provided context.

\subsection{Evaluation of Dataset}
\label{sec:evaluation_dataset}

\paragraph{Evaluation Framework}
We assess the MI quality and general quality of KMI based on human evaluation, particularly by experts.
We recruited four Korean psychological counseling experts\footnote{They are professional counseling psychologists certified by the Korean Psychological Association.} experienced in MI.
For evaluation, we randomly sample 100 dialogues from each of KMI, CounselGPT, and AnnoMI. Sampling details of KMI are available in Appendix~\ref{sec:KMI_sampling}. The experts are then requested to evaluate the dialogues based on the criteria defined in Section \ref{sec:evaluation_criteria}, using a Likert scale ranging from 1 to 5. The median score is calculated by applying the majority vote approach. The evaluation form we used is shown in Figure \ref{fig:eval_form} of Appendix \ref{sec:eval_form}.

\paragraph{MI Quality}
As demonstrated in Table \ref{tab:evaluation_dataset}, KMI surpasses CounselGPT and AnnoMI across all evaluation criteria of MI quality. CounselGPT, which is not built on MI theory, shows mostly low scores across the MI quality criteria. AnnoMI achieves high scores on the MI quality criteria but shows a lower similarity score than our dataset.
This might be due to the quality degradation that occurs during translation, as a subtle difference in nuance could be significant in the context of psychotherapy.
Finally, achieving a score of 3.94 in similarity and 3.74 in effectiveness illustrates that our dataset qualifies as a valuable psychotherapeutic resource.

\paragraph{General Quality}
As shown in Table \ref{tab:evaluation_dataset}, KMI exhibits a high score of 4.65 in consistency and 4.22 in fluency, both of which outperform CounselGPT and AnnoMI. This indicates that our framework, tailored to Korean, capably generates high-quality dialogues.
Also, the on-topic score of 4.17 demonstrates that the dialogues in KMI showcase a strong alignment with actual Korean situations. As CounselGPT and AnnoMI do not contain context data, on-topic is not evaluated on these datasets.

\begin{table}[t!]
    \small
    \begin{center}
    \setlength{\tabcolsep}{6pt} %
    \renewcommand{\arraystretch}{0.95}
    \resizebox{\columnwidth}{!}{%
    \begin{tabular}{@{}lrrr@{}}
    \toprule
    \hspace{1.5mm}\textbf{Dataset} & KMI & CounselGPT & AnnoMI \\ \midrule
    \hspace{2mm}\textbf{MI Quality}       &      &      &       \\
    \hspace{4mm}Partnership   & \textbf{4.40} & 2.78 & 4.03  \\
    \hspace{4mm}Acceptance    & \textbf{4.35} & 2.86 & 3.99  \\
    \hspace{4mm}Compassion    & \textbf{3.81} & 3.05 & 3.50  \\
    \hspace{4mm}Evocation     & \textbf{4.07} & 2.88 & 3.87  \\
    \hspace{4mm}Similarity    & \textbf{3.94} & 2.96 & 3.72  \\
    \hspace{4mm}Effectiveness & \textbf{3.74} & 2.66 & 3.45  \\ \midrule
    \hspace{2mm}\textbf{General Quality}  &      &      &       \\
    \hspace{4mm}Consistency   & \textbf{4.65} & 3.69 & 3.82  \\
    \hspace{4mm}Fluency       & \textbf{4.22} & 3.40 & 2.25  \\
    \hspace{4mm}On-Topic      & \textbf{4.17} &   -  &   -   \\ \bottomrule
    \end{tabular}%
    }
    \end{center}
    \caption{Human evaluation results of MI quality and general quality for sampled dialogues from KMI, CounselGPT, and AnnoMI. All differences in pairwise comparisons between KMI and the other two datasets are statistically significant at a significance level of 0.01.}
    \label{tab:evaluation_dataset}
\end{table}

\paragraph{MI Label Accuracy}
In addition to the qualitative evaluation, we measure the accuracy of MI labels in the dataset to validate its reliability. We randomly sample 30 utterances for each MI label from KMI, resulting in a total of 210 utterances\footnote{\textit{Other} is excluded.}. Sampling details are available in Appendix~\ref{sec:MI_sampling}. We then ask the experts to assess whether each utterance matches its assigned label, using \texttt{True/False} evaluation. Results in Table~\ref{tab:acc-labels} demonstrate that most of the utterances are consistent with their label, with an average accuracy of 96.0\%.

\begin{table}[t!]
    \begin{center}
    \renewcommand{\arraystretch}{0.95}
    \resizebox{\columnwidth}{!}{%
    \begin{tabular}{p{5cm}>{\raggedleft\arraybackslash}p{2.5cm}}
    \toprule
    \hspace{1mm}\textbf{MI Label}   & \textbf{Accuracy (\%)} \\ \midrule
    \hspace{1mm}Simple Reflection  & 96.7\hspace{1mm}          \\
    \hspace{1mm}Complex Reflection & 96.7\hspace{1mm}          \\
    \hspace{1mm}Open Question      & 100.0\hspace{1mm}         \\
    \hspace{1mm}Closed Question    & 95.0\hspace{1mm}          \\
    \hspace{1mm}Affirm             & 96.7\hspace{1mm}          \\
    \hspace{1mm}Give Information   & 90.0\hspace{1mm}          \\
    \hspace{1mm}Advise             & 96.7\hspace{1mm}          \\ \midrule
    \hspace{1mm}Average            & \textbf{96.0}\hspace{1mm} \\ \bottomrule
    \end{tabular}%
    }
    \end{center}
    \caption{Accuracy of the correspondence between utterances and MI labels.}
    \label{tab:acc-labels}
\end{table}

\subsection{Evaluation of Dialogue Model}
\label{sec:evaluation_chatbot}

\paragraph{Training Dialogue Models}
We assess dialogue models fine-tuned with these datasets to verify the utility of KMI as a training dataset for chatbots. 
We fine-tune \texttt{komt-llama2-7b-v1}\footnote{\url{https://github.com/davidkim205/komt}}, an LLaMA 2-7B~\citep{touvron2023llama} model further tuned with Korean multi-task instruction tuning, using each of KMI, CounselGPT, and translated AnnoMI.
Fine-tuning details are stated in Appendix \ref{sec:train_chatbot}.

\paragraph{Evaluation Framework}
We recruited four native Korean crowdworkers to participate in conversations with the dialogue models. Given context data, they are asked to interactively converse with the model as if they were the writer of the given context. They engage in 30 conversations with each dialogue model, using an identical set of 30 contexts for each model. Among the collected data, context data not previously used for data generation is used in this step.
The completed dialogues are then evaluated by experts\footnote{The same experts as in Section \ref{sec:evaluation_dataset}.} based on the aforementioned criteria, using a 5-point Likert scale.

\begin{table}[t!]
    \small
    \begin{center}
    \setlength{\tabcolsep}{6pt} %
    \renewcommand{\arraystretch}{0.95}
    \resizebox{\columnwidth}{!}{%
    \begin{tabular}{@{}lrrr@{}}
    \toprule
    \hspace{1.5mm}\textbf{Dialogue Model} & KMI & CounselGPT & AnnoMI \\ \midrule
    \hspace{2mm}\textbf{MI Quality}       &      &      &       \\
    \hspace{4mm}Partnership   & \textbf{4.43} & 2.97 & 3.17  \\
    \hspace{4mm}Acceptance    & \textbf{4.04} & 2.69 & 2.74  \\
    \hspace{4mm}Compassion    & \textbf{3.29} & 2.31 & 2.44  \\
    \hspace{4mm}Evocation     & \textbf{3.70} & 2.92 & 2.73  \\
    \hspace{4mm}Similarity    & \textbf{3.18} & 2.60 & 2.46  \\
    \hspace{4mm}Effectiveness & \textbf{3.00} & 2.10 & 2.28  \\ \midrule
    \hspace{2mm}\textbf{General Quality}  &      &      &       \\
    \hspace{4mm}Consistency   & \textbf{4.30} & 3.70 & 3.40  \\
    \hspace{4mm}Fluency       & \textbf{3.55} & 2.70 & 2.50  \\ \bottomrule
    \end{tabular}%
    }
    \end{center}
    \caption{Human evaluation results of MI quality and general quality for dialogue models fine-tuned using KMI, CounselGPT, and AnnoMI. All differences in pairwise comparisons between KMI and the other two datasets are statistically significant at a significance level of 0.01.}
    \label{tab:evaluation_chatbot}
\end{table}

\paragraph{MI Quality}
Results in Table~\ref{tab:evaluation_chatbot} show that the dialogue model fine-tuned on KMI outperforms other models across all MI quality criteria by a substantial margin. This highlights the value of KMI for building mental health chatbots, especially those specialized in MI.
We speculate that the relatively low scores for similarity and effectiveness are related to the capability of the base model, \texttt{komt-llama2-7b-v1}, since it might be challenging for a 7B-sized model to proficiently role-play a professional therapist or lead an effective MI session from a professional perspective.
The scores of CounselGPT and AnnoMI are mostly below 3, implying limitations in their suitability for training MI chatbots.

As shown in dialogue examples from each model in Appendix~\ref{sec:dialogue_examples}, the model fine-tuned with KMI generates utterances that comply with the principles of MI. In contrast, models fine-tuned with CounselGPT and AnnoMI lack the features of MI and tend to produce excessively lengthy utterances.

\paragraph{General Quality}
Results in Table~\ref{tab:evaluation_chatbot} also show that the dialogue model fine-tuned on KMI demonstrates better consistency and fluency than other models.
AnnoMI displays the lowest scores in these metrics, indicating the limitations of using a transcribed dataset for chatbot development.
On-topic is not evaluated in dialogue models because it's the user’s role to raise and discuss content related to the context data, not the model's.

\section{Discussion and Conclusion}
\label{sec:conclusion}
In this paper, we introduced a novel framework to generate synthetic motivational interviewing dialogues, along with KMI, the resulting dataset of 1,000 Korean MI dialogues. Through comprehensive evaluations, we demonstrated its quality and utility for chatbot development. For meaningful evaluation, we also proposed novel metrics derived from MI theory.

Our dataset has three main applications. 
First, it could be used for developing mental health chatbots grounded in MI theory. We trained a Korean chatbot in this paper, where expert evaluation results and usage examples (see Table \ref{tab:chatbot-example-kmi}) show that our dataset is capable of building effective chatbots.
Second, it could function as a labeled dataset of \texttt{(Utterance, MI label)} pairs for classification and forecasting tasks. MI label accuracy of 96.0\% demonstrates the reliability of the labels.
Third, it could serve as a reference for MI practitioners. Full dialogues of therapy sessions are usually private, while MI textbooks provide only segments of dialogues. Despite being synthetic, KMI can offer high-quality, full dialogues covering diverse topics.

In addition, our generation method is potentially generalizable to other languages. Korean-specific resources required in our framework are: (1) context data, (2) 3\textasciitilde4 high-quality examples for each MI label for few-shot prompting, and (3) a Korean-to-English translation model. If these resources are available, our framework could be further used to boost non-English resources for psychotherapy.

\section*{Limitations}
\label{sec:limitations}
\paragraph{Error Analysis} 
While our framework generally performs well, we found some cases where the dialogue doesn't sufficiently reflect the contents and details of the context data. These cases are discussed below:

(1) 
When the client simulator generates an utterance, it considers both the context and dialogue history. The ideal scenario is when it responds to the therapist while incorporating the details of the context at the same time, but sometimes the counselor might lead the conversation in a direction that doesn't align with the context data. In such cases, the client simulator primarily focuses on responding to the therapist's last utterance, potentially resulting in somewhat shallow or meandering dialogues that don't contain the details of the context data.

(2)
As the context data is collected from an online counseling platform, some of it contains extreme or violent content. We found that GPT-4 automatically filters such content when generating utterances, removing problematic elements from the final KMI dataset. Dialogues generated based on such context might be less specific and realistic.

\paragraph{MI Forecaster Trained on English Dataset}
As there were no public Korean MI resources available prior to our work, we used AnnoMI, an English MI dialogue dataset, to train the MI forecaster. AnnoMI consists entirely of English dialogues, most of which are based on Western cultures. Thus, the predictions made by the MI forecaster might be biased toward the practices of therapists from these cultural backgrounds. However, we suppose that the core principles of MI such as partnership, acceptance, compassion, and evocation remain universal regardless of the language and culture. 
It is known that MI is well-suited to working with different languages and culturally diverse populations \citep{miller2023motivational}.
Also, \citet{mcmaster2015motivational} discovered that over 50 groups of health professionals across North America, Europe, Africa, and Asia generated remarkably similar responses when asked what they would consider good practice and bad practice in their own settings. This indicates that good MI practice would still remain good MI practice in different cultures, supporting the validity of our approach.

\paragraph{Effect of Translation}
Two kinds of translations are involved in our research. First, in the generation process of KMI, we translate the generated dialogue history into English to use it as the input for the MI forecaster. Second, we translate AnnoMI into Korean during the evaluation phase\footnote{Since KMI is the first Korean dataset grounded in MI, we had to choose an MI dataset from a different language for comparison.}. In both cases, we utilized Upstage SOLAR \citep{kim-etal-2024-solar}, one of the most recent translation models, to preserve the meaning of sentences, but some degradation is inevitable.
We consider the latter more critical since slight variations in nuance or expression can have a substantial impact in the context of psychotherapy, possibly resulting in lower scores for AnnoMI.
We believe the effect of translation is more trivial in the former case, as the MI forecaster considers the overall content and flow of the dialogue history to make predictions, rather than focusing on detailed expressions.

\section*{Ethics Statement}
\label{sec:ethical-considerations}
\paragraph{Dataset Curation} In the Korean psychological counseling platform \textit{Mindcafe}, individuals are anonymous and they recognize that their writing is archived on the site unless they delete it. We collected only publicly available data and there was no interaction with the Mindcafe users.

\paragraph{Human Evaluation} For human evaluation, we recruited four professional counselors. They were compensated with 50,000 won per hour. For the evaluation of dialogue models, we also recruited four crowdworkers. They were compensated with 10,000 won per hour, which is higher than the Korean minimum wage at the time they worked.

\paragraph{Mental Health Support Chatbots}
The KMI dataset was designed to facilitate the advancement of chatbots that adhere to MI principles, particularly for mental health support. However, despite the promising capabilities of recent AI models, AI-driven chatbots may still pose risks. The unpredictability of generative models can lead to unintended consequences, especially in sensitive conversations involving emotional distress.
Therefore, any deployment of chatbots trained on KMI should be approached with caution, and human supervision is essential to ensure that the responses are appropriate.

\section*{Acknowledgments}
This work was supported by the BK21 FOUR Program (Education and Research Center for Industrial Innovation Analytics) funded by the Ministry of Education, Korea (No. 4120240214912) and the National Research Foundation of Korea (NRF) grant funded by the Korea government (MSIT) (No. 2021R1A2C2093785, RS-2024-00333484, RS-2024-00414981).

\bibliography{anthology,custom}

\clearpage
\appendix

\section{Context Data Details} %
\label{sec:context_data}
We collect a total of 7,530 posts in seven categories from Mindcafe. Table~\ref{tab:collection-details} shows the number of collected data by category and the number of data that received 3-point scores. The number and ratio of data by score are presented in Table~\ref{tab:per-score-result}. Scoring is conducted using few-shot prompting with GPT-4 API, and the prompt used is shown in Table~\ref{tab:prompt-for-scoring}. The filtered context data are then randomly sampled with predetermined quantities for each category and used for data generation. The number of data sampled for each category is detailed in Table~\ref{tab:context-data}.

\begin{table}[h!]
    \small
    \begin{center}
    \renewcommand{\arraystretch}{1.2}
    \begin{tabular}{@{}lrr@{}}
    \toprule
    \hspace{2mm}\textbf{Category}           & \textbf{\# Data}\hspace{2mm} 
                                            & \textbf{\# Filtered Data}\hspace{2mm} \\ \midrule
    \hspace{2mm}Mental Health               & 2,651\hspace{2mm} & 1,102\hspace{2mm}  \\
    \hspace{2mm}Interpersonal Relationships & 1,338\hspace{2mm} & 535\hspace{2mm} \\
    \hspace{2mm}Ego \& Personality          & 1,300\hspace{2mm} & 538\hspace{2mm} \\
    \hspace{2mm}Career \& Employment        & 874\hspace{2mm}   & 477\hspace{2mm} \\
    \hspace{2mm}Academic \& Examination     & 645\hspace{2mm}   & 247\hspace{2mm} \\
    \hspace{2mm}Family                      & 527\hspace{2mm}   & 126\hspace{2mm} \\
    \hspace{2mm}Addiction \& Obsession      & 195\hspace{2mm}   & 73\hspace{2mm} \\
    \midrule
    \hspace{2mm}Total                       & 7,530\hspace{2mm} & 3,098\hspace{2mm}  \\
    \bottomrule
    \end{tabular}
    \end{center}
    \caption{Number of data collected by category and filtered by a score of 3 within each category.}
    \label{tab:collection-details}
\end{table}

\begin{table}[h!]
    \small
    \begin{center}
    \setlength{\tabcolsep}{15pt}
    \renewcommand{\arraystretch}{1.2}
    \begin{tabular}{@{}crr@{}}
    \toprule
    \hspace{2mm}\textbf{Score}  & \textbf{\# Data} & \textbf{\%}\hspace{2mm} \\ \midrule
    \hspace{2mm}3     & 3,603   & 47.9\hspace{2mm} \\
    \hspace{2mm}2     & 3,729   & 49.5\hspace{2mm} \\
    \hspace{2mm}1     & 198     & 2.6\hspace{2mm}  \\ \midrule
    \hspace{2mm}Total & 7,530   &      \\ \bottomrule
    \end{tabular}
    \end{center}
    \caption{Number and percentage of data by score.}
    \label{tab:per-score-result}
\end{table}

\begin{table}[h!]
    \small
    \begin{center}
    \renewcommand{\arraystretch}{1.2}
    \begin{tabular}{@{}lr@{}}
    \toprule
    \hspace{2mm}\textbf{Category}           & \textbf{\# Data}\hspace{2mm} \\ \midrule
    \hspace{2mm}Mental Health               & 200\hspace{2mm}  \\
    \hspace{2mm}Interpersonal Relationships & 200\hspace{2mm}  \\
    \hspace{2mm}Ego \& Personality          & 200\hspace{2mm}  \\
    \hspace{2mm}Career \& Employment        & 200\hspace{2mm}  \\
    \hspace{2mm}Academic \& Examination     & 100\hspace{2mm}  \\
    \hspace{2mm}Addiction \& Obsession      & 50\hspace{2mm}   \\
    \hspace{2mm}Family                      & 50\hspace{2mm}   \\ \midrule
    \hspace{2mm}Total                       & 1,000\hspace{2mm}   \\
    \bottomrule
    \end{tabular}
    \end{center}
    \caption{Number of context data used to generate data by category.}
    \label{tab:context-data}
\end{table}

\begin{table}[h!]
    \small
    \begin{center}
    \begin{tabularx}{1.0\linewidth}{@{}X@{}}
    \toprule
    \textbf{Instruction:} \\
    Below is an utterance from a client who has come for counseling with a therapist. Based on the provided definition of Motivational Interviewing (MI), the characteristics of topics suitable for MI, examples of appropriate and inappropriate utterances, and the classification criteria, please classify whether the utterance below is appropriate or not. Do not explain the reason, just answer with a number only. \\ \midrule
    \textbf{Definition of MI:} \\
    It is a cooperative and goal-oriented conversation method that strengthens an individual's own motivation and commitment to change. Set within an accepting and empathetic atmosphere, it encourages individuals to explore and articulate their own reasons for change, thereby bolstering their personal motivation and commitment toward a specific goal. \\ \midrule
    \textbf{Characteristics of Topics Suitable for MI:} \\
    - Cases where an individual wants to change but struggles due to ambivalence about the change or lack of motivation. \\
    - Problems that can be somewhat improved if the individual decides to make an effort on their own.\\ \midrule
    \textbf{Example of Appropriate Utterance:} \\
    Lately, I've been having these thoughts. My heart is complex, and my mind is troubled, making me feel exhausted lately and like I've lost the vitality I used to have. I just don't seem to have any energy, I feel lethargic, and everything seems to be my fault. It's really hard. I'm very afraid, and simply living feels exhausting. These days, I just want to take sleeping pills and sleep for two days without thinking about anything. Why is this happening? What should I do in times like these? \\ \midrule
    \textbf{Example of Inappropriate Utterance:} \\
    I've been having a lot of strange dreams. In my dreams, a woman in red clothes tries to kill me with a knife, and things like that. I've actually seen ghosts and heard their voices. It really feels like someone is constantly watching me... My grandmother says that young people shouldn't go to shamans and that it's not an option for me. I'm really at a loss; I'm so scared even now, feeling restless... I don't want to go to a psychiatrist, and I can't go to a shaman... What should I do? \\ \midrule
    \textbf{Scoring Criteria:} \\
    - 3 points: The content is specific, and the topic is suitable for MI. \\
    - 2 points: The content is specific, but the topic is not suitable for MI. \\
    - 1 point: The content is not specific or has some flaw; counseling about someone other than oneself (family, friends, etc.) scores 1 point\\ \midrule
    \textbf{Utterance:} \\
    I'm not sure where things went wrong, but my life seems to have been fraught with problems right from the start. Interacting with others, including my family, and even dealing with myself, is a challenge. Starting anything feels like a monumental task; where others glide through effortlessly, I find myself having to exert many times more effort just to keep up. I'm always having to plan several steps ahead for anything I do, and navigating relationships and society feels like an extension of this perpetual struggle. Everything feels so hard. Giving up has become a part of my daily routine. How can I turn my situation around? \\ \bottomrule
    \end{tabularx}
    \end{center}
    \caption{Prompt for scoring the collected context data.}
    \label{tab:prompt-for-scoring}
\end{table}

\clearpage

\section{Preprocessing AnnoMI}
\label{sec:annomi_preprocessing}
We follow the steps below to preprocess the AnnoMI dataset for our task.

(1) As we want to simulate high-quality MI sessions, we only use the dialogues annotated \texttt{high-quality} in the original dataset, which leaves 110 dialogues out of 133.

(2) For therapist utterances that have several MI labels, we only leave the one annotated as \texttt{main\_therapist\_behaviour} in the original dataset, so that each therapist utterance has only one label.

(3) For the six MI labels shown in Table \ref{tab:label_integration} that correspond one-to-one in AnnoMI and our taxonomy, we directly use them with slight modifications in the label names.

(4) For the therapist utterances not categorized as one of the labels in Table \ref{tab:label_integration}, we further classify them using the RoBERTa-based MI classifier developed in \citet{welivita2023boosting}. We add \textit{Affirm} label to the utterances the classifier predicts as \textit{Affirm}.

(5) The remaining therapist utterances are labeled as \textit{Other}.

\begin{table}[h!]
    \small
    \begin{center}
    \renewcommand{\arraystretch}{1.05}
    \begin{tabular}{c c}
    \toprule
    \textbf{AnnoMI Label} &  \textbf{Our Label} \\ \midrule
    reflection\_simple     & Simple Reflection   \\
    reflection\_complex    & Complex Reflection   \\
    question\_open         & Open Question   \\
    question\_closed       & Closed Question   \\
    input\_information     & Give Information   \\
    input\_advice          & Advise   \\ \bottomrule
    \end{tabular}
    \end{center}
    \caption{MI labels that correspond one-to-one in AnnoMI and our taxonomy.}
    \label{tab:label_integration}
\end{table}

\newpage

\section{Details of MI Forecaster}
\label{sec:mi_forecaster_details}

We show the number of converted data for training the MI forecaster according to its history window size in Table \ref{tab:t5_dataset_num}. If the input text exceeds the maximum token length of 512, we truncate it from the left side to preserve the recent history.
An example of the converted data is shown in Table \ref{tab:t5_dataset_ex}.

Hyperparameters for training the MI forecaster are listed in Table \ref{tab:hyperparameters}. We use AdamW \citep{loshchilov2017decoupled} for optimization.

\begin{table}[h!]
    \small
    \begin{center}
    \renewcommand{\arraystretch}{1}
    \begin{tabular}{@{}c r@{}}
    \toprule
    \hspace{2mm}\textbf{History Window Size}  &  \textbf{\# Data}\hspace{2mm} \\ \midrule
    \hspace{2mm}1        & 4,346\hspace{2mm}   \\
    \hspace{2mm}2        & 4,329\hspace{2mm}   \\
    \hspace{2mm}3        & 4,236\hspace{2mm}   \\
    \hspace{2mm}4        & 4,219\hspace{2mm}   \\
    \hspace{2mm}5        & 4,126\hspace{2mm}   \\
    \hspace{2mm}6        & 4,109\hspace{2mm}   \\
    \hspace{2mm}7        & 4,016\hspace{2mm}   \\
    \hspace{2mm}8        & 3,999\hspace{2mm}   \\ \bottomrule
    \end{tabular}
    \end{center}
    \caption{Number of data for training the MI forecaster.}
    \label{tab:t5_dataset_num}
\end{table}

\begin{table}[h!]
    \small
    \begin{center}
    \setlength{\tabcolsep}{4pt} %
    \renewcommand{\arraystretch}{1.2}
    \begin{tabular}{p{1.5cm}p{5.7cm}} %
    \toprule
    \multicolumn{2}{c}{\textbf{With Therapist Label}} \\ 
    \midrule
    Input Text & Predict the next therapist's dialogue act: \textbf{[Therapist: Open Question]} Uh, what else can you tell me about your drinking? \textbf{[Client]} Well, I usually drink when I'm at home trying to unwind and I drink while I'm watching a movie. And sometimes, um, I take a bath but I also drink when I take a bath sometimes. \\
    Output Text & \textbf{[Therapist: Open Question]} \\
    \midrule
    \multicolumn{2}{c}{\textbf{Without Therapist Label}} \\ 
    \midrule
    Input Text & Predict the next therapist's dialogue act: \textbf{[Therapist]} Uh, what else can you tell me about your drinking? \textbf{[Client]} Well, I usually drink when I'm at home trying to unwind and I drink while I'm watching a movie. And sometimes, um, I take a bath but I also drink when I take a bath sometimes. \\
    Output Text & \textbf{[Therapist: Open Question]} \\
    \bottomrule
    \end{tabular}
    \end{center}
    \caption{An example of the dataset used to train MI forecaster (window size = 2).}
    \label{tab:t5_dataset_ex}
\end{table}

\begin{table}[h!]
    \small
    \begin{center}
    \renewcommand{\arraystretch}{1}
    \begin{tabular}{@{}l r@{}}
    \toprule
    \hspace{2mm}\textbf{Hyperparameter}  &  \textbf{Value}\hspace{2mm} \\ \midrule
    \hspace{2mm}Epochs                 & 5\hspace{2mm}   \\
    \hspace{2mm}Learning Rate          & 1e-4\hspace{2mm}   \\
    \hspace{2mm}Batch Size             & 8\hspace{2mm}   \\
    \hspace{2mm}Max. Sequence Length   & 512\hspace{2mm}   \\ \bottomrule
    \end{tabular}
    \end{center}
    \caption{Hyperparameters for training the MI forecaster.}
    \label{tab:hyperparameters}
\end{table}

\section{Change Talk}
\label{sec:change_talk}
Change talk is a client language that indicates movement toward a particular change \citep{miller2023motivational}. Four types of change talk—desire, ability, reasons, and need (often abbreviated as DARN)—are called preparatory change talk because you hear them when people are considering whether to do something. Descriptions and examples of each type are stated below.

\begin{itemize}
    \item \textbf{Desire language} is a way of saying, “I want.” It signals some inclination toward action. \\
    (“I \textit{want} to quit smoking.”)
    \item \textbf{Ability language} provides information about how confident people are that they would be able to take the action in question. \\
    (“I think it’s \textit{possible} for me to quit.”)
    \item \textbf{Reason language} states specific reasons for doing something. The reason might be a possible advantage of change or a disadvantage of not changing. \\
    (“My children are begging me to quit.”)
    \item \textbf{Need language} has an imperative quality emphasizing some urgency of change. It implies that a change is important without specifying why. \\
    (“I’ve \textit{got} to quit smoking.”)
\end{itemize}

\clearpage
\onecolumn
\section{Prompt Templates}
\label{sec:prompt_templates}

\begin{table*}[hbt]
    \begin{center}
    \small
    \renewcommand{\arraystretch}{1.4}
    \begin{tabularx}{1.0\linewidth}{@{}X@{}}
    \toprule
    \textbf{Instruction:} \\
    당신은 상담사(counselor)입니다. 아래의 제약조건을 지키고, 주어진 대화 다음에 올 상담사(counselor)의 발화(utterance)를 만드세요. 대화를 통해 내담자가 현재 상황에서 변화하고자 하는 내적 동기를 이끌어내는 것이 이 상담의 목적입니다. 이 때, '단순 반영하기(Simple Reflection)'에 근거해서 생성해주세요.\\
    \midrule
    \textbf{Constraints:} \\
    - 존댓말을 사용해야 합니다.\\
    - 답변은 1문장 또는 2문장으로 작성해야 합니다.\\
    - 한 문장은 한 절(clause)로만 구성해주세요. `\textasciitilde며' 등의 연결어를 사용하지 말아주세요.\\
    - 상대방을 칭할 때 `내담자님'이라고 말해주세요.\\ 
    \midrule
    \textbf{Definitions and Examples:} \\
    Definition of `Simple Reflection':\\
    내담자가 한 말에 정보를 더 추가하기보다는 내담자가 한 말에 가장 가깝게 반영하는 것을 말한다.\\
    질문의 형태보다 진술문의 형태로 반영하는 것이 더 좋다.\\
    내담자가 말한 내용 중 핵심만을 반복한다.\\\\
    Example of `Simple Reflection' \#1:\\
    내담자: 제 여동생처럼만 되었으면 좋겠어요.\\
    상담사 [단순 반영하기]: 동생처럼 되고 싶으시군요\\
    \\
    Example of `Simple Reflection' \#2:\\
    내담자: 생각해 보세요. 엄마도 제가 연락하지 않으면 전화 한 통 안 해요. 물론 돈이 필요할 땐 연락을 하시죠. 그러니 엄마에게 뭘 기대하겠어요.\\
    상담사 [단순 반영하기]: 엄마한테 기대할 게 없다는 말씀이시군요.\\
    ...\\
    \midrule
    \textbf{Dialogue History:} \\
    ...\\
    \bottomrule
    \end{tabularx}
    \end{center}
    \caption{Prompt template for therapist utterance generation.}
    \label{tab:prompt_table_1}
\end{table*}

\begin{table*}[ht]
    \begin{center}
    \small
    \renewcommand{\arraystretch}{1.4}
    \begin{tabularx}{1.0\linewidth}{@{}X@{}}
    \toprule
    \textbf{Instruction:} \\
    당신은 아래 상황의 문제로 상담사(counselor)에게 상담하러 간 내담자(client)입니다. 아래의 제약조건을 지키고, 주어진 대화 다음에 올 내담자(client)의 답변(response)을 만드세요. 아래 상황에 대한 고민을 상담사(counselor)에게 천천히, 조금씩 이야기해보세요. 만약 상담사와의 대화로 인해 이 상황을 변화시키고 싶은 의지가 생겼고 대화 맥락 상 변화 대화를 하는 것이 자연스러울 경우 `변화 대화'를 생성해주세요.\\
    \midrule
    \textbf{Constraints:} \\
    - 존댓말을 사용해야 합니다.\\
    - 답변은 1문장 또는 2문장으로 작성해야 합니다.\\
    - 한 문장은 한 절(clause)로만 구성해주세요. `\textasciitilde며' 등의 연결어를 사용하지 말아주세요.\\
    - 주어진 상황을 바탕으로 하되, 구체적인 상황을 자연스럽게 지어내서 추가해도 됩니다.\\
    - 생성하는 내담자의 답변이 이전 대화와 자연스럽게 연결되어야 합니다.\\
    \midrule
    \textbf{Definitions and Examples:} \\
    Definition of `Change Talk':\\
    자신의 문제에 대해 스스로 말하기를 변화하고 싶다거나 변화와 관련된 진술 혹은 언어표현이다.\\
    아래와 같은 것들이 변화 대화에 포함된다:\\
    1. 변화에 대한 희망, 변화하고 싶다는 언어적 진술과 표현 (Desire)\\
    2. 변화할 수 있다는 생각, 변화에 대한 낙관적인 시각, 변화는 가능하다 혹은 변화할 것이라는 표현 (Ability)\\
    3. 변화의 이득과 장점, 변화로 인해 긍정적인 결과가 생길 것이라는 표현 (Reason)\\
    4. 변화의 필요성, 변화하지 않을 때의 문제점과 손실, 변화하지 않는 것에 대한 걱정, 염려 및 우려 (Need)\\\\
    Example of `Change Talk' \#1:\\
    내담자: 사람들이랑 더 많은 대화를 나누게 된다면 더 많은 친구들을 사귈 수 있을 거에요.\\
    \\
    Example of `Change Talk' \#2:\\
    내담자: 이렇게 계속 살이 찌면 안 돼요.\\
    ...\\
    \midrule
    \textbf{Context:} \\
    ...\\
    
    \midrule
    \textbf{Dialogue History:} \\
    ...\\
    \bottomrule
    \end{tabularx}
    \end{center}
    \caption{Prompt template for client utterance generation.}
    \label{tab:prompt_table_2}
\end{table*}

\clearpage
\twocolumn
\section{Evaluation Criteria}
\label{sec:eval_criteria}

\paragraph{MI Quality}
In our work, the following criteria are used to evaluate MI quality:
\begin{itemize}
    \item{\textbf{Partnership}} assesses whether the therapist respects the client's expertise, defined as the client's experiences with the issues and related experiences they encounter, and maintains a balanced power dynamic, fostering mutual collaboration.
    \item{\textbf{Acceptance}} evaluates whether the therapist accepts the client as an autonomous individual, acknowledging their capacity for imperfection and mistakes without judgment.
    \item{\textbf{Compassion}} assesses the therapist's mindset based on deep understanding and empathy towards the client, aimed at alleviating their difficulties.
    \item{\textbf{Evocation}} assesses the therapist's capability to evoke the client's intrinsic resources, such as values, life goals, strengths, personality traits, and motivations.
    \item\textbf{{Similarity}} measures how closely the generated therapist's utterances resemble those of an actual therapist.
    \item{\textbf{Effectiveness}} measures the overall efficacy of the MI session.
\end{itemize}

\paragraph{General Quality}
The following criteria are used to evaluate general quality:
\begin{itemize}
    \item{\textbf{Consistency}} assesses whether the utterances with each turn are consistent and if the overall flow of the dialogue is maintained consistently.
    \item{\textbf{Fluency}} evaluates whether the therapist and client's utterances in the dialogue are natural and fluent and whether the overall flow of the dialogue maintains a natural progression.
    \item{\textbf{On-Topic}} measures whether the dialogue remains relevant to the provided context and continues in accordance with the context provided.
\end{itemize}

\newpage

\section{Training Dialogue Models}
\label{sec:train_chatbot}

\paragraph{Fine-tuning Details}
We fine-tune the \texttt{komt-llama2-7b-v1} model for KMI, CounselGPT, and AnnoMI by using 3 NVIDIA RTX A6000 GPUs with 40GB memory. Table~\ref{tab:hyperparameters-dial} illustrates hyperparameters for fine-tuning the dialogue models.
We use AdamW for optimization.

\begin{table}[hbt!]
    \small
    \begin{center}
    \renewcommand{\arraystretch}{1.05}
    \begin{tabular}{@{}l r@{}}
    \toprule
    \hspace{2mm}\textbf{Hyperparameter}  & \textbf{Value}\hspace{2mm} \\ \midrule
    \hspace{2mm}Epochs                   & 3\hspace{2mm}      \\
    \hspace{2mm}Learning Rate            & 1e-6\hspace{2mm}   \\
    \hspace{2mm}Batch Size               & 16\hspace{2mm}     \\
    \hspace{2mm}Max. Sequence Length  & 2,048\hspace{2mm}    \\ \bottomrule
    \end{tabular}
    \end{center}
    \caption{Hyperparameters for fine-tuning the dialogue models.}
    \label{tab:hyperparameters-dial}
\end{table}

\section{Sampling Details}
\label{sec:sampling}

\subsection{Sampling Details of KMI}
\label{sec:KMI_sampling}
Table~\ref{tab:sampling-kmi} indicates the number of dialogues sampled from each category for evaluation purposes. A total of 100 dialogues are sampled.

\begin{table}[hbt!]
    \small
    \begin{center}
    \renewcommand{\arraystretch}{1}
    \begin{tabular}{@{}lr@{}}
    \toprule
    \hspace{2mm}\textbf{Category}           & \textbf{\# Data}\hspace{2mm} \\ \midrule
    \hspace{2mm}Mental Health               & 16\hspace{2mm}  \\
    \hspace{2mm}Interpersonal Relationships & 14\hspace{2mm}  \\
    \hspace{2mm}Ego \& Personality          & 14\hspace{2mm}  \\
    \hspace{2mm}Career \& Employment        & 14\hspace{2mm}  \\
    \hspace{2mm}Academic \& Examination     & 14\hspace{2mm}  \\
    \hspace{2mm}Addiction \& Obsession      & 14\hspace{2mm}   \\
    \hspace{2mm}Family                      & 14\hspace{2mm}   \\ \midrule
    \hspace{2mm}Total                       & 100\hspace{2mm}   \\
    \bottomrule
    \end{tabular}
    \end{center}
    \caption{Number of dialogues sampled by category.}
    \label{tab:sampling-kmi}
\end{table}

\FloatBarrier

\subsection{Sampling Details of MI Labels}
\label{sec:MI_sampling}
We randomly sample 30 utterances for each of the seven MI labels from the generated dialogues. As the number of dialogues differs by category, efforts were made to ensure that categories were distributed as evenly as possible within the utterances sampled for each MI label. Table~\ref{tab:mi-label-result} shows the number of sampled utterances for each MI label and category.

\begin{table*}[h!]
    \small
    \renewcommand{\arraystretch}{1.2}
    \begin{tabular}{@{}lrrrrrrr@{}}
    \toprule
    \textbf{Category}           & \multicolumn{1}{c}{\textbf{\begin{tabular}[c]{@{}c@{}}Simple\\ Reflection\end{tabular}}} & \multicolumn{1}{c}{\textbf{\begin{tabular}[c]{@{}c@{}}Complex\\ Reflection\end{tabular}}} & \multicolumn{1}{c}{\textbf{\begin{tabular}[c]{@{}c@{}}Open\\ Question\end{tabular}}} & \multicolumn{1}{c}{\textbf{\begin{tabular}[c]{@{}c@{}}Closed\\ Question\end{tabular}}} & \multicolumn{1}{c}{\textbf{Affirm}} & \multicolumn{1}{c}{\textbf{\begin{tabular}[c]{@{}c@{}}Give\\ Information\end{tabular}}} & \multicolumn{1}{c}{\textbf{Advise}} \\ \midrule
    Mental Health               &  6  &  6  &  6  &  6  &  6  &  6  &  2\hspace{2mm}  \\
    Interpersonal Relationships &  4  &  4  &  4  &  5  &  4  &  5  &  5\hspace{2mm}  \\
    Ego \& Personality          &  4  &  4  &  4  &  4  &  4  &  5  & 10\hspace{2mm}  \\
    Career \& Employment        &  4  &  4  &  4  &  4  &  4  &  5  &  9\hspace{2mm}  \\
    Academic \& Examination     &  4  &  4  &  4  &  4  &  4  &  5  &  2\hspace{2mm}  \\
    Addiction \& Obsession      &  4  &  4  &  4  &  4  &  4  &  4  &  1\hspace{2mm}  \\
    Family                      &  4  &  4  &  4  &  3  &  4  &  -  &  1\hspace{2mm}  \\ \midrule
    Total                       &  30 &  30 &  30 &  30 &  30 &  30 &  30\hspace{2mm} \\ \bottomrule
    \end{tabular}
    \caption{Number of utterances sampled by MI label and category.}
    \label{tab:mi-label-result}
\end{table*}

\clearpage
\onecolumn

\section{Evaluation Form}
\label{sec:eval_form}

\begin{figure*}[h!]
    \centering 
    \includegraphics[width=0.95\linewidth]{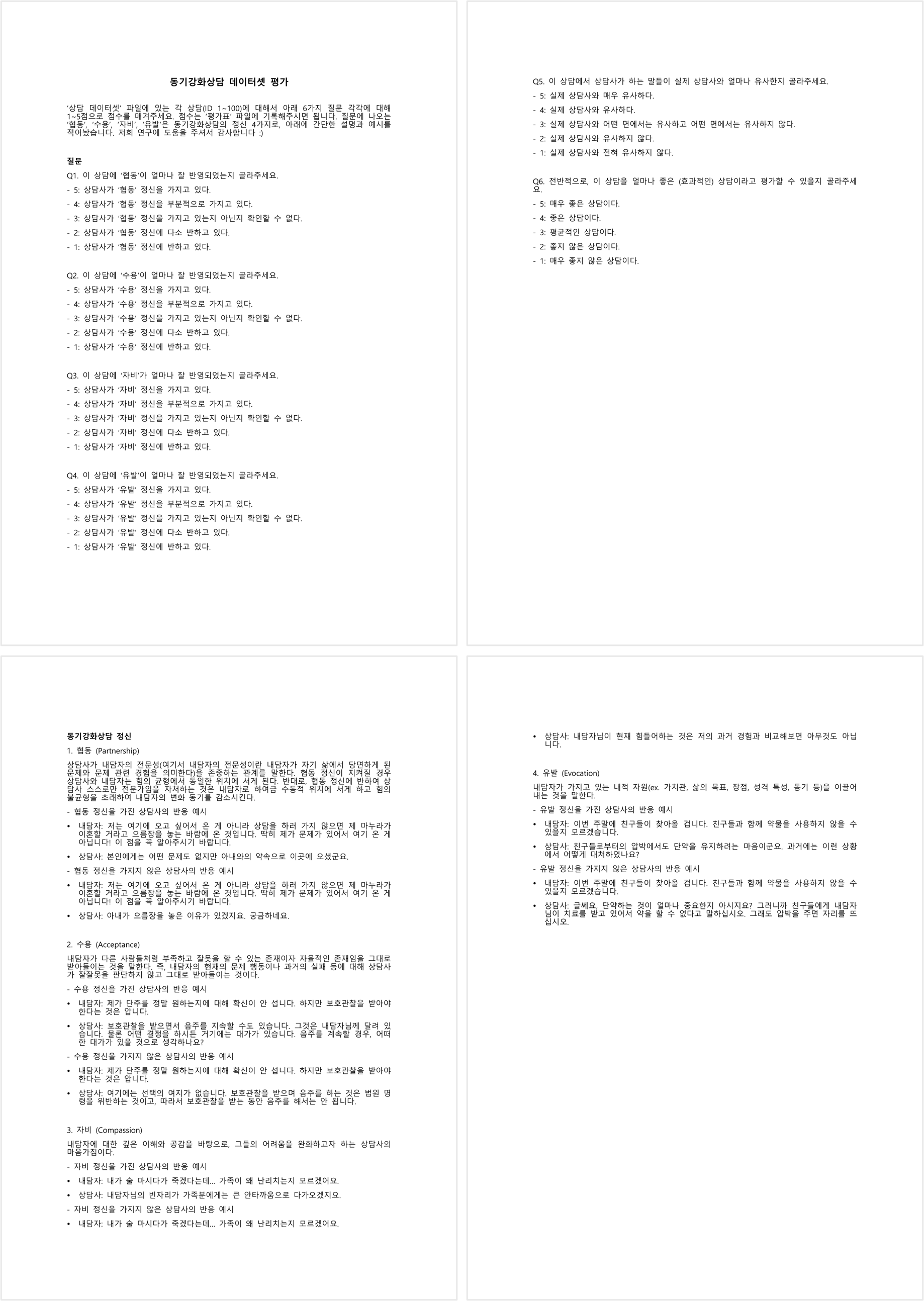}
    \caption{Instructions and questions related to each metric for expert evaluation of MI quality. For reference, we also provided descriptions of each metric, along with examples of good and bad responses for each metric.} 
    \label{fig:eval_form}
\end{figure*}

\clearpage

\section{Examples of Dialogues in KMI}
\label{sec:dialogue_examples_kmi}

\begin{table*}[hbt!]
    \begin{center}
    \small
    \renewcommand{\arraystretch}{1.3}
    \begin{tabular}{@{}p{0.08\linewidth}p{0.69\linewidth}p{0.18\linewidth}@{}}
    \toprule
    \textbf{Role} & \textbf{Utterance} & \textbf{MI Label} \\
    \midrule
    Therapist &안녕하세요, 최근 어떤 일이나 사건이 내담자님의 생각이나 기분에 영향을 주었나요?& [Open Question] \\
    &Hello, have there been any recent events or incidents that have affected your thoughts or mood?&\\
    Client & 고3 때 겪은 입시 실패가 제 마음에 큰 상처를 줬어요.& - \\
    &The failure of my college entrance exams when I was a senior in high school left a big scar on my heart.&\\
    Therapist &그 시기의 경험이 지금의 내담자님에게 아직도 큰 영향을 미치고 있군요.  & [Complex Reflection] \\
     &The experience of that time still has a big impact on you now.&\\
    Client & 네, 그때의 실패 이후로 저는 더 이상 그림을 그리는 것에서 행복을 느끼지 못하고 있어요.  & - \\
    &Yes, I don't feel happy from drawing anymore after that failure.&\\
    Therapist & 그림을 그릴 때 어떤 감정이나 생각이 드시나요?& [Open Question] \\
    &What emotions or thoughts do you have when you paint?&\\
    Client & 이젠 그냥 무기력하달까, 예전처럼 그림에 대한 열정이나 동기가 전혀 생기지 않아요. & - \\
    &Now I'm just feeling kind of lethargic, I don't have any passion or motivation for drawing like I used to.&\\
    Therapist & 그렇군요, 과거의 실패로 인해 그림에 대한 열정과 동기가 상실되었네요.& [Complex Reflection] \\
    &I see, your passion and motivation for painting has been lost due to past failures.&\\
    Client & 맞아요, 그리고 제 생활 패턴도 엉망이 됐어요. & -\\
    &Right, and my life pattern is messed up too.&\\
    Therapist & 과거의 경험이 생활 패턴에도 부정적인 영향을 주고 있는 상황이시군요.& [Complex Reflection] \\
    &I see, it seems like your past experiences are also negatively affecting your lifestyle patterns.&\\
    Client & 네, 제가 진짜 원하는 게 뭔지, 하고 싶은 게 뭔지도 잘 모르겠어요. & - \\
    &Yeah, I don't even know what I really want, or what I want to do.&\\
    Therapist & 하고 싶은 것이 무엇인지 잘 모르겠다는 말씀이시군요.& [Simple Reflection] \\
    &I see, you're not sure what you want to do.&\\
    Client & 네, 저 자체가 행동력이 없어진 것 같아서 앞으로 나아갈 방향을 찾기가 어려워요. & - \\
    &Yeah, I feel like I've lost my momentum and it's hard to find a direction to move forward.&\\
    Therapist & 앞으로 나아가기 위해 어떤 것들이 필요하다고 생각하시나요?& [Open Question] \\
    &What do you think are the things that are needed to move forward?&\\
    Client & 감정을 다시 찾고, 무엇을 하고 싶은지 결정하는데 도움이 될 방향성이 필요해요. & - \\
    &I need to find my emotions again, and I need some direction to help me decide what I want to do.&\\
    Therapist & 감정을 찾고, 하고 싶은 것을 결정하는 과정에서 어떤 경험이나 활동이 도움이 될 것 같으신가요?& [Open Question] \\
    &What experiences or activities do you think would be helpful in the process of finding your feelings and deciding what you want to do?&\\
    Client & 그림과 관련된 취미 활동을 조금씩 다시 해보는 것이 좋을 것 같아요. & -\\
    &I think I should start doing some of my hobby activities related to drawing again.&\\
    Therapist & 그림과 관련된 취미 활동을 다시 시작해보실 생각이시군요.& [Simple Reflection] \\
    &I see, you're thinking about picking up a hobby related to painting again.&\\
    \bottomrule
    \end{tabular}
    \end{center}
    \caption{
    First example of a full dialogue from KMI.}
    \label{tab:KMIFullExample1}
\end{table*}

\begin{table*}[hbt!]
    \begin{center}
    \small
    \renewcommand{\arraystretch}{1.3}
    \begin{tabular}{@{}p{0.08\linewidth}p{0.69\linewidth}p{0.18\linewidth}@{}}
    \toprule
    \textbf{Role} & \textbf{Utterance} & \textbf{MI Label} \\
    \midrule
    Therapist &안녕하세요, 최근 어려움을 겪고 있는 부분이 있으신가요?& [Open Question] \\
    &Hello, is there anything you've been struggling with lately?&\\
    Client &네, 저는 가족과 관계가 안 좋아지고 나서 스스로에 대한 혐오가 커져가고 있어요. & - \\
    &Yes, I've been feeling more and more disgust for myself after my relationship with my family went bad.&\\
    Therapist &그러셨군요. 가족과의 문제가 자신에 대한 느낌에도 영향을 끼치고 있군요.& [Complex Reflection] \\
    &I see. Your issues with your family are affecting how you feel about yourself.&\\
    Client &네, 특히 어머니와의 관계가 가장 힘들어요. & - \\
    &Yes, my relationship with my mother is the hardest.&\\
    Therapist &어머니와의 관계에서 어떤 부분이 가장 힘드신가요?& [Open Question] \\
    &What part of your relationship with your mother is the most difficult for you?&\\
    Client &어머니가 저에게 실망하셨다고 말씀하시는 것과 제가 무슨 행동을 해도 따뜻한 말 한마디 듣지 못하는 것입니다. & - \\
    &My mother telling me she is disappointed in me and no matter what I do I never get a warm word.&\\
    Therapist &어머니로부터 듣고 싶은 말씀이 어떤 것인지 여쭤보아도 될까요?& [Open Question] \\
    &May I ask what you would like to hear from your mother?&\\
    Client &그냥 ``잘했다", ``수고했다" 같은 말을 듣고 싶어요. & - \\
    &I just want to hear ``good job", ``well done" and stuff like that.&\\
    Therapist &따뜻한 격려의 말을 듣고 싶으신군요.& [Simple Reflection] \\
    &You would like to hear some warm words of encouragement.&\\
    Client &네, 어머니의 짧은 격려 한마디가 저에게 큰 힘이 될 것 같아요. & - \\
    &Yes, I think a short word of encouragement from my mother would be very helpful for me.&\\
    Therapist &그런 격려의 말을 듣게 되면 어떤 느낌이 드실 것 같으세요?& [Open Question] \\
    &How would you feel if you heard those encouraging words?&\\
    Client &그럼 조금이나마 제가 하는 일에 대해 인정받는 기분이 들어서 더 열심히 하고 싶어질 것 같아요. & - \\
    &Then I'll feel a little bit more appreciated for what I do and want to do it more.&\\
    Therapist &어머니의 격려가 내담자님의 자신감을 회복시키고 더욱 분발하는 데 필요한 원동력이 되시겠군요.& [Complex Reflection] \\
    &It sounds like your mother's encouragement boosts you to regain your confidence and push yourself further.&\\
    Client &네, 정말 그렇게 된다면 저도 다시 희망을 가지고 더 노력할 수 있을 것 같아요. & - \\
    &Yes, I think that would really give me hope again and I would try harder.&\\
    Therapist &어머니와의 관계에서 이러한 변화를 이루기 위해 시도해보고 싶으신 방법이 있으신가요?& [Open Question] \\
    &Is there anything you would like to try to make these changes in your relationship with your mother?&\\
    Client &어머니께 제가 느끼는 이 마음을 솔직히 털어놓고 어떻게 하면 관계를 개선할 수 있는지 이야기해보고 싶어요. & - \\
    &I want to be honest with my mother about how I feel and discuss how we can improve our relationship.&\\
    Therapist &어머니와 솔직한 대화를 나누며 관계 개선을 위한 첫 발을 내딛고자 하는 모습이 보기 좋군요.& [Affirm] \\
    &It's great that you're willing to have an honest conversation with your mother and take the first step toward improving your relationship.&\\
    \bottomrule
    \end{tabular}
    \end{center}
    \caption{Second example of a full dialogue from KMI.}
    \label{tab:KMIFullExample2}
\end{table*}

\begin{table*}[hbt!]
    \begin{center}
    \small
    \renewcommand{\arraystretch}{1.3}
    \begin{tabular}{@{}p{0.08\linewidth}p{0.69\linewidth}p{0.18\linewidth}@{}}
    \toprule
    \textbf{Role} & \textbf{Utterance} & \textbf{MI Label} \\
    \midrule
    Therapist &안녕하세요, 최근에 어떤 일들이 마음에 걸리게 되셨나요?& [Open Question] \\
    &Hi, what things have been on your mind lately?&\\
    Client & 회사에서 직책을 맡은 지 3년이 되었는데, 주변 사람들이 점점 떠나가고 있어요. 처음에는 업무가 바빠 크게 신경을 못 썼는데, 최근에 와서 생각해보니 제가 뭘 잘못했는지, 마음이 많이 무겁습니다.& - \\
    &I've been in my position at my company for three years, and people around me are leaving individually. At first, I was busy with work and didn't pay much attention, but recently, I've been thinking about what I did wrong, and it's been weighing on my mind.&\\
    Therapist &그 상황에서 자신의 역할과 주변 사람들의 변화에 대해 고민하게 되셨군요.& [Simple Reflection] \\
    &I see, it sounds like you're struggling with your role in that situation and the changes in the people around you.&\\
    Client &네, 그런데 사람들이 떠나가는 이유가 제가 잘못한 것인지, 아니면 다른 이유 때문인지 헷갈려요. & - \\
    &Yes, but I'm confused if the reason people are leaving is because of something I did, or if it's something else.&\\
    Therapist &내담자님의 내부에서는 책임감과 혼란스러움이 공존하고 있네요.& [Complex Reflection] \\
    &There is a sense of responsibility and confusion within you.&\\
    Client & 맞아요, 그래서 저를 응원해주시는 분들의 말도 가끔 의심하게 되고 경계하게 됩니다.& - \\
    &Right, so I also sometimes doubt and become wary of the words of those cheering me on.&\\
    Therapist &응원하는 분들의 말씀조차도 믿기 어려운 상태가 되셨군요.& [Complex Reflection] \\
    &Even the words of well-wishers are hard to believe.&\\
    Client &그렇습니다. 결국 SNS 앱까지 삭제했어요, 남들 행복한 모습을 보는 게 부담스럽게 느껴져서요. & - \\
    &I did. I eventually deleted my SNS apps, too, because I felt burdened seeing others' happy moments.&\\
    Therapist &SNS 앱 삭제는 부담감을 줄이려는 내담자님의 방법이었군요.& [Simple Reflection] \\
    &Deleting SNS apps was a way for the client to reduce their burden.&\\
    Client &네, 그런데 앱을 삭제해도 마음이 가벼워지진 않아요. & - \\
    &Yes, but deleting the app doesn't make me feel any lighter.&\\
    Therapist &앱을 삭제한 것이 마음의 부담을 완전히 해소시키지는 못했네요.& [Complex Reflection] \\
    &I guess deleting the app didn't completely relieve your mind.&\\
    Client &네, 앱을 삭제했지만 현실의 문제는 여전히 남아있어요. & - \\
    &Yes, I deleted the app, but the real-life problem still remains.&\\
    Therapist &SNS 앱을 삭제하는 과정에서 내담자님의 해결을 향한 노력이 느껴지지만, 그 배후에는 여전히 해결되지 않은 현실의 문제들이 존재하는 것 같네요.& [Complex Reflection] \\
    &It seems that in deleting SNS apps, the client's efforts towards resolution are evident, but the underlying issues remain unresolved.&\\
    Client &네, 정말 그렇습니다. 어디서부터 어떻게 시작해야 할지 막막합니다. & - \\
    &Yes, exactly. I'm at a loss as to where and how to start.&\\
    Therapist &이 상황을 개선하기 위해 내담자님이 가장 먼저 해보고 싶은 일이 무엇인지 궁금합니다.& [Open Question] \\
    &I'm curious to know what you would like to do first to improve this situation.&\\
    Client &주변 동료들과의 관계를 개선해보고 싶어요. & - \\
    &I want to improve my relationships with my colleagues around me.&\\
    Therapist &동료들과의 관계 개선에 초점을 맞추고 싶어하시는군요.& [Simple Reflection] \\
    &I see, you want to focus on improving your relationships with your colleagues.&\\
    \bottomrule
    \end{tabular}
    \end{center}
    \caption{Third example of a full dialogue from KMI.}
    \label{tab:KMIFullExample3}
\end{table*}

\clearpage

\section{Examples of Dialogues with Chatbots Fine-tuned on Each Dataset}
\label{sec:dialogue_examples}

\begin{table*}[hbt!]
    \begin{center}
    \small
    \renewcommand{\arraystretch}{1.2}
    \begin{tabular}{p{1.6cm}p{13.2cm}}
    \toprule
    \textbf{Role} & \textbf{Utterance} \\ \midrule
    Therapist     & 안녕하세요, 최근 어떤 일들이 마음에 무거움을 주고 있나요?\\ 
    &Hello, may I inquire about any recent events causing a sense of heaviness in your heart? \\
    Client        & 어떤 일을 하기 전에 귀찮다는 생각이 먼저 들어서 마음 먹기가 어려워요.\\ 
    &\begin{tabular}[c]{@{}l@{}}Before beginning a task, I frequently encounter the notion that it will be bothersome, which makes it \\challenging to prepare myself mentally.\end{tabular}\\
    Therapist     & 그 일들이 마음에 무거움을 주고 계신 것 같군요. 그 일들을 하기 전에 먼저 귀찮아하는 생각이 먼저\\
    &들어서 마음 먹기가 어려운 것 같군요.\\
    &\begin{tabular}[c]{@{}l@{}}It seems that those tasks have weighed heavily on your mind. It appears you find it difficult to get\\mentally prepared because the thought of them being bothersome comes first before even starting.\end{tabular} \\
    Client        & 네 맞아요. 사소한 것 조차 실행하기가 어렵게 느껴지고 무기력해지는 것 같아요.\\
    &Yes, that's correct. Even minor tasks feel difficult to execute, and I find myself feeling listless.\\
    Therapist     & 사소한 일들이 당장 너무 큰 문제로 느껴지고 무기력해지는 것 같군요.\\ 
    &It seems that even minor tasks feel overwhelmingly problematic at the moment, leading to a sense of\\
    &listlessness.\\
    Client        & 네. 그래서 점점 더 예전 만큼 사기가 오르지도 않고 그냥 그러려니 하루를 보내게 돼요.\\ 
    &Yes. As a result, I find myself less motivated than before, just letting the days pass by with a sense of\\
    &resignation.\\
    Therapist     & 그렇게 되면서 큰 문제라고 느끼고 계시는군요. 그런 순간에 처음에는 어떻게 생각하시는 것 같으신가요?\\
    &It seems that this situation has become a significant concern for you. What were your initial thoughts during such moments?\\
    Client        & 처음에는 뭔가 해보려고 했는데 점점 혼자 있고 싶어지고 스스로가 답답해져요.\\
    &Initially, I tried to do something about it, but gradually, I found myself wanting to be alone and feeling frustrated with myself.\\
    Therapist     & 스스로 답답해지는 과정에서 주로 어떤 감정을 느끼시나요?\\
    &What kind of emotions do you primarily feel when becoming frustrated with yourself?\\
    Client        & 이런 것들을 다른 사람들에게 숨기는 것도 어려워져서 저도 이제 변하고 싶어요.\\
    &It's becoming harder to hide these things from others, and now I really want to change.\\
    Therapist     & 그 변화에 대한 당신의 열망이 인상적입니다. 시작하기 위해 작은 변화로 무엇을 시도해보고싶은지 궁금해요.\\
    &Your desire for change is impressive. I am interested in knowing what small change you would like to try as a start.\\
    \bottomrule
    \end{tabular}%
    \end{center}
    \caption{Example of a dialogue with a chatbot fine-tuned on KMI.}
    \label{tab:chatbot-example-kmi}
\end{table*}

\begin{table*}[hbt!]
    \begin{center}
    \small
    \renewcommand{\arraystretch}{1.25}
    \begin{tabular}{p{1.6cm}p{13.2cm}}
    \toprule
    \textbf{Role} & \textbf{Utterance} \\ \midrule
    Therapist    & 안녕하세요, 현재 어떤 감정을 가장 많이 느끼고 계신가요?\\ 
    &Hello, what emotion are you feeling the most right now? \\
    Client       & 진짜 펑펑 소리내면서 울면서 속 좀 풀고 싶은데 눈물이 안나요.\\ 
    &I really want to cry my heart out and relieve my feelings, but the tears just won't come.\\
    Therapist    & 언제나 그런 건 잘 알고 있으신가요? 그런 감정은 매우 편안하고 안정되면서도 조금 불안한 설렘을\\
    &느끼는 것 같아요. 이런 감정은 매우 흔하게 나타나며, 대부분 긍정적인 상황에서 자주 느끼곤 합니다.\\
    &그러나 어떤 부분에서는 언제나 불안한 기분을 느끼기도 하시는 거 같아요.이런 불안한 기분은 일상\\
    &적인 상황에서 자주 느껴지는 것 같아요.\\
    &Are you always aware of this? It seems like such emotions bring a mix of comfort and stability, along\\
    &with a slight, anxious excitement. Such feelings are quite common and typically arise in predominantly\\
    &positive contexts. Nonetheless, it seems there is always an aspect of you that remains uneasy. This feeling\\
    &of uneasiness seems to be a frequent occurrence in your daily life.\\
    Client      & 원래도 남들에게 제 얘기도 잘 안하고, 힘든거 내색하는 스타일도 아니라서 타인에게 위로 받는 것도\\
    &익숙하지 않다보니 답답한 마음 털어 놓을 곳도 없어요.\\
    &I've always been the type to keep personal matters to myself and not show when I'm struggling. As a\\
    &result, I'm not accustomed to receiving comfort from others, leaving me nowhere to release my frustrations.\\
    Therapist   & 그렇군요. 그러나 자신에게 위로 받는 것은 매우 중요합니다. 그러니 자신의 감정에 대해 얘기해보는\\
    &것이 좋겠어요. 그러면 당신의 감정에 대해 얘기하면서 어떤 상황에서 불안하게 느껴지는지 자세히\\
    &알려주세요. 그리고 그 상황에서 어떤 방식으로 대처하고 있는지 알려주세요. 그러면 당신의 감정에\\
    &대한 이해도가 높아질 것입니다.\\ 
    &I see. However, receiving comfort for oneself is crucial. So, it would be good to talk about your emotions.\\
    &Could you tell me more about when you feel anxious and how you're dealing with it in those situations?\\
    &By sharing your emotions, we can better understand them.\\
    Client      & 그냥 아무에게도 말 안하고 끙끙 앓고있어요. 어떻게 극복할 수 있을까요?\\ 
    &I'm just keeping it all to myself and suffering in silence. How can I overcome this?\\
    Therapist   & 그렇군요. 이러한 불안한 감정은 자신에게 불안감을 주는 원인이 될 수 있습니다. 이러한 감정을 억제\\
    &하는 방법을 찾아보는 것이 좋습니다. 이러한 방법은 자신에게 맞는 방법을 찾는 것이 중요합니다.\\
    &자신에게 맞는 방법을 찾기 위해 여러 가지 방법을 시도해보세요.\\
    &Indeed, such anxious feelings can become a source of anxiety themselves. It's advisable to look for ways\\
    &to suppress these emotions. It's important to find a method that suits you. Try various methods to find the\\
    &one that fits you best.\\
    Client        & 어떤 방법들이 좋을까요?\\
    &What methods might be good?\\
    Therapist   & 당신의 감정에 대해 이야기하고 불안한 감정에 대해 얘기해보았습니다. 이러한 대화는 당신의 감정에\\
    &대한 이해도가 높아질 것입니다. 당신이 감정에 대해 이야기하고 불안한 감정에 대해 이야기하면서\\
    &어떤 방식으로 대처하고 있는지 알려주세요.\\
    &We've talked about your emotions and feelings of anxiety. Such conversations will increase your understanding of your emotions. Please share how you are dealing with these feelings as you talk about your\\
    &emotions and discuss your anxious feelings.\\ \bottomrule
    \end{tabular}
    \end{center}
    \caption{Example of a dialogue with a chatbot fine-tuned on CounselGPT.}
    \label{tab:chatbot-example-counselgpt}
\end{table*}

\begin{table*}[hbt]
    \begin{center}
    \small
    \renewcommand{\arraystretch}{1.25}
    \begin{tabular}{p{1.6cm}p{13.2cm}}
    \toprule
    \textbf{Role} & \textbf{Utterance} \\ \midrule
    Therapist    & 안녕하세요, 최근 어떤 상황이나 생각이 마음에 걸리게 되었나요?\\ 
    &Hello, have there been any recent situations or thoughts that have been weighing on your mind?\\
    Client       & 중요한 결정들을 너무 쉽게 저질러버리고 인내심이나 책임감이 너무 없는 것 같아요. 이런 점을\\
    &어떻게 고칠 수 있을까요?\\ 
    &I often make important decisions too quickly and feel that I lack patience and responsibility. How can I\\
    &work on improving in these areas?\\
    Therapist    & 당신의 상황에 대해 더 자세히 알려주시면, 도움을 드릴 수 있을 것 같아요. 결정을 내리는 데 도움이\\
    &될 수 있는 다른 방법이 있을까요?\\
    &If you can share more details about what you're going through, I might be able to help. Are there other\\
    &approaches that could assist you in making decisions?\\
    Client      & 휴학이나 퇴사 같은 인생의 중요한 결정을 할때 뚜렷한 목표나 계획없이 단순히 인간관계가 마음에\\
    &안 든다는 이유로 하곤 했어요.\\
    &When making significant life decisions, like taking a leave of absence from school or quitting a job, I often\\
    &did so without a clear goal or plan, simply because I was dissatisfied with my interpersonal\\
    &relationships.\\
    Therapist   & 이런 상황을 극복하기 위해서는 몇 가지 방법을 고려해볼 수 있어요. 이러한 상황에 대해 알아보고\\
    &싶어요. 첫째로, 결정을 내릴 때 중요한 것은 단지 단기적인 것이 아니라 장기적인 것이라고 생각해\\
    &봅시다. 결정을 내릴 때 너무 쉽게 저질러버리지 않도록 잊지 말아야 합니다.\\ 
    &To overcome such situations, we can consider a few methods. I'm interested in exploring this situation\\
    &situation further. First off, when making decisions, it's important to consider not just the short-term\\
    &but also the long-term implications. It's crucial not to rush into decisions too hastily.\\
    Client   & 맞아요. 근데 제가 너무 현실감각이 없는 것 같아요. 어렸을 때부터 하고 싶은 것만 하려고 했거든요.\\ 
    &That's right. But I feel like I lack a sense of reality. I've always tended to just do what I wanted, ever since\\
    &I was a kid.\\
    Therapist   & 네, 그건 중요한 문제가 아닙니다. 이런 경우에는 몇 가지 조언을 드릴게요. 첫째로, 너무 극단적인\\
    &선택을 하지 마세요. 결정을 내리기 전에 충분한 정보를 수집하고, 다른 옵션들을 고려해보세요.\\
    &결정을 내리기 전에 너무 압도적으로 생각하는 것은 결정을 내리는 데 좋지 않을 수 있습니다.\\
    &Yes, that's not a major issue. In cases like this, I'd like to offer some advice. First, try not to make extreme\\
    &choices. Make sure to gather enough information before making a decision, and consider all other options.\\
    &Overthinking things before deciding can actually hinder you from making a sound decision.\\
    Client        & 저도 그렇게 하려고 하는데 성격이 급해서 계속 저질러 버리게 되는 것 같아요.\\
    &I try to do that, but I think my impulsive nature keeps leading me to rush into things.\\
    Therapist  & 네, 그것은 심리적인 문제일 수 있습니다. 이런 경우에는 적절한 조언을 드리기 위해 전문가의 도움을\\
    &받는 것이 좋을 수 있습니다. 마지막으로, 자신의 능력과 자원을 인정하는 것이 중요합니다. 자신에게\\
    &자신감을 북돋아 주는 것은 결정을 내리는 데 도움이 될 것입니다. 자신에게 맞는 결정을 내리는 것이\\
    &중요한 것이랍니다.\\
    &Yes, it could be a psychological issue. In such situations, seeking help from a professional for proper\\
    &guidance might be helpful. Also, acknowledging your own abilities and resources is important. Building\\
    &up your self-confidence can really help when it comes to making decisions. It's essential to make choices\\
    &that are truly right for you.\\ \bottomrule
    \end{tabular}
    \end{center}
    \caption{Example of a dialogue with a chatbot fine-tuned on AnnoMI.}
    \label{tab:chatbot-example-annomi}
\end{table*}

\end{document}